\documentclass[twoside,11pt]{article}
\usepackage[utf8]{inputenc}
\usepackage{longtable} 
\usepackage{textcomp} 
\usepackage{ragged2e} 
\usepackage[a4paper, margin=1in]{geometry} 
\usepackage{array}
\usepackage{blindtext}
\usepackage{longtable} 
\usepackage{comment}   
\usepackage{multirow}
\usepackage{booktabs}       
\usepackage{amsmath} 
\newtheorem{proof}{Proof}

\usepackage{dmlr2e}

\newcommand{\acc}{\mbox{acc}}
\newcommand{\cov}{\mbox{cov}}

\usepackage{lastpage}

\ShortHeadings{LxCIM}{T. Brogueira and M. Figueiredo}
\firstpageno{1}

\begin{document}

\title{LxCIM: a new rank-based binary classifier performance metric invariant to local exchange of classes}

\author{\name Tiago Brogueira \email tiago.brogueira@tecnico.ulisboa.pt \\
       \addr Instituto de Telecomunicações\\
       Instituto Superior Técnico\\
       Lisboa, Portugal
       \AND
       \name Mário Figueiredo \email mario.figueiredo@tecnico.ulisboa.pt \\
       \addr Instituto de Telecomunicações\\
       Instituto Superior Técnico\\
       Lisboa, Portugal}


\maketitle

\begin{abstract}
Binary classification is one of the oldest, most prevalent and studied problems in machine learning. However, the metrics used to evaluate model performance have received comparatively little attention. The \textit{area under the receiver operating characteristic curve} (AUROC) has long been a standard choice for model comparison. Despite its advantages, AUROC is not always ideal, particularly for problems that are invariant to \textit{local exchange of classes} (LxC), a new form of metric invariance introduced in this work.
To address this limitation, we propose LxCIM (\textit{LxC-invariant metric}), which is not only rank-based and invariant under local exchange of classes, but also intuitive, logically consistent, and always computable, while enabling more detailed analysis through the cumulative accuracy–decision rate curve. Moreover, LxCIM exhibits clear theoretical connections to AUROC, accuracy, and the area under the accuracy–decision rate curve (AUDRC). These relationships allow for multiple complementary interpretations: as a symmetric form of AUROC, a rank-based analogue of accuracy, or a more representative and more interpretable variant of AUDRC.
Finally, we demonstrate the direct applicability of LxCIM to the bivariate causal discovery problem (which exhibits invariance to local exchange of classes) and show how it addresses the acknowledged limitations of existing metrics used in this field. All code and implementation details are publicly available at \url{https://github.com/tiagobrogueira/Causal-Discovery-In-Exchangeable-Data}.
\end{abstract}


\section{Introduction}
Binary classification is the task of assigning an item, based on its features (or attributes), to one of two mutually exclusive classes, often conventionally referred to as positive and negative. It constitutes one of the most fundamental and widely studied problems in machine learning, with myriad applications such as medical diagnosis, credit scoring, quality control, and spam detection \citep{binaryclassificationdef}. 

\subsection{Landscape of Binary Classification Metrics}
Evaluation metrics to assess binary classifiers can be grouped into three categories based on the type of model output they evaluate \citep{separationmetrics}. The simplest category relies purely on the model’s categorical prediction. A more informative category, named rank-based metrics, evaluates the score assigned by the model, which is assumed to carry ordinal meaning: higher scores indicate a higher likelihood that an input belongs to one class. (Categorical predictions are typically derived from these scores by applying a decision threshold.) Finally, the most refined category interprets the score as a calibrated (posterior) probability of one of the classes, in the interval $[0,1]$; this naturally sets the optimal decision threshold at $0.5$ under a 0/1 loss.  

Probabilistic outputs provide the richest information about a model’s predictive performance, and metrics that leverage these probabilities yield more accurate and informative assessments of model quality \citep{separationmetrics}. However, interpreting outputs probabilistically imposes stricter constraints on the class of models that can be implemented, limiting flexibility in certain applications. Consequently, in this work, the focus will be on evaluation metrics for models that output a score, but it will not be assumed that the score has a probabilistic interpretation, thus excluding probabilistic metrics. 

For categorical metrics, the \textit{confusion matrix} provides a concise representation of all four possible outcomes of a categorical classifier’s predictions on labeled data: \textit{true positives} ($\text{TP}$), \textit{false positives} ($\text{FP}$), \textit{false negatives} ($\text{FN}$), and \textit{true negatives} ($\text{TN}$) \citep{buildingconfusion}. When models output continuous scores that are subsequently thresholded (the focus of this work), the confusion matrix captures the concrete decision outcomes associated with that particular threshold. From these four quantities ($\text{TP}$, $\text{FP}$, $\text{FN}$, and $\text{TN}$), a variety of other well-known standard performance measures can be derived, including \textit{accuracy, sensitivity, specificity}, and \textit{precision}. Other metrics, such as the F1-score, the \textit{false discovery rate} (FDR), or the \textit{Matthews correlation coefficient} (MCC), can be expressed as functions of these same counts \citep{benchmetrics}. 

Finally, rank-based metrics assess the model’s outputs based on the relative ordering of the predicted scores. Specifically, the model’s scores are interpreted as having an ordinal (though not probabilistic) meaning: higher scores indicates a greater likelihood of belonging to the positive class. Consequently, these predicted scores are first transformed into their corresponding ranks within the evaluation set, giving rise to the term \textit{rank-based} \citep{ksstatistic}. The primary motivation behind rank-based approaches is that purely categorical metrics disregard the model’s confidence in individual predictions. In such metrics, each sample's contribution is independent of the model's confidence in it (there is no distinction between misclassifying an instance with a 0.51 score, versus one with a 0.90 score \citep{separationmetrics}. By evaluating the relative ordering of predicted scores, rank-based metrics enable comparisons across different predictions, thereby providing a measure of performance that reflects the model’s ability to discriminate between instances with varying levels of confidence.

Another notable property of rank-based metrics is their invariance under any monotonic (order-preserving) transformation of the predicted scores. Thus, these metrics are unaffected by normalization or other calibration procedures. However, this invariance comes with the limitation that their results can be difficult to match to the actual decisions made by the model. Therefore, a threshold choice is often required for proper interpretability \citep{rankingchoice1}. A further limitation is that computing rank-based metrics typically requires access to the entire evaluation set, as the ranking operation is inherently non-modular and cannot be computed incrementally.

\label{auroc_intro}
The most relevant rank-based metric is the \textit{area under the receiver operating characteristic}\footnote{The expression ``receiver operating characteristic" comes from its initial use in analyzing receivers for radar signals during World War II.} curve, commonly referred to as AUROC, AUC ROC, or even simply AUC \citep{buildingconfusion}. The ROC curve plots the true positive rate $ \text{TPR} = \text{TP} / (\text{TP} + \text{FN} )$, which represents the proportion of correctly predicted positive instances, as a function of the false positive rate $ \text{FPR} =  \text{FP} /( \text{FP} + \text{TN})$, which represents the proportion of negative instances incorrectly predicted as positive. Alternatively, the AUROC can be interpreted probabilistically as $\mathbb{P}[s(X^+) > s(X^-)]$, that is, the probability that a randomly chosen positive instance $X^+$ is assigned a higher score than a randomly chosen negative one $X^-$. 

The ROC curve can be obtained by sweeping the decision threshold across the range of predicted scores. At one extreme, all predictions are negative, yielding $\text{FPR} = 0$ and $\text{TPR} = 0$; at the other extreme, all predictions are positive, thus $\text{FPR} = 1$ and $\text{TPR} = 1$. In the ideal scenario, all positive instances receive higher scores than all negative instances, resulting in a curve that forms a square passing through the point $(\text{FPR} = 0, \text{TPR} = 1)$ and corresponding to $\text{AUROC} = 1$. This means there is a threshold for which all positive and negative instances are classified correctly. Conversely, for a random classifier, the true positive rate equals the false positive rate for all thresholds, thus $\text{TPR} = \text{FPR}$, yielding $\text{AUROC} = 0.5$.

There are other rank-based metrics, such as the area under the precision-recall curve (AUPRC), the Kolmogorov–Smirnov statistic, which can be interpreted as the maximum difference between the TPR and FPR integrals \citep{ksstatistic}, and the area under the convex hull of the ROC curve (AUCH) \citep{areaunderconvexhull}, corresponding to the area under the circle defined by the worst $(\text{FPR}, \text{TPR})$ point in the ROC curve, i.e., satisfying $\text{TPR}^2 + (1 - \text{FPR})^2$. Nonetheless, all these metrics fundamentally follow the same procedure as AUROC, sweeping across the decision threshold and computing alternative statistics based on the confusion matrix. Consequently, they share many of the strengths and limitations of the AUROC.

The last alternative approach to evaluating binary classifiers considers the possibility that a model may abstain from making a prediction.  This is relevant in domains where an incorrect decision carries a significantly higher cost than deferring a decision, such as medical diagnosis, making it crucial to assess performance across varying decision rates. To accommodate this, the score function is complemented with a decision function, which produces a value intended to reflect the model's confidence \citep{auacc}. The following proposition (proved in Appendix \ref{decision_function_appendix}) gives conditions for a function to qualify as a decision function. 
\begin{proposition} \label{prop:bimono}
For a function $g(s)$ to qualify as a decision function, it must be bi-monotonic: strictly decreasing for all $s < s^*$ and strictly increasing for all $s > s^*$. 
\end{proposition}

This structure implicitly encodes a fixed decision threshold at $s^*$, thereby delineating which examples are predicted positive and which are negative.

\citet{rejectionrate} propose analysing accuracy-decision rate curves to assess the model performance when constrained to making only a certain fraction of their most confident predictions. This enables a more detailed analysis, while allowing the model to be tuned to meet a desired minimum performance. 

Building on this idea, \citet{audrc} propose the \textit{area under the accuracy-decision rate curve} (AUDRC) defined as
\begin{equation}    
\text{AUDRC} = \frac{1}{N} \sum_{i=1}^{N} \frac{1}{i}
\sum_{j=1}^{i} \mathbb{I}\left[y_{\pi(j)} = u(s_{\pi(
j)})\right],
\label{AUDRC}
\end{equation}
\noindent where $\pi$ denotes the index permutation that orders the samples such that $g(s_{\pi(1)}) \ge g(s_{\pi(2)}) \ge \dots \ge g(s_{\pi(N)})$, where $g(s)$ is the bi-monotonic decision function, $N$ represents the total number of examples, $y_j$ denotes the label of the $j$-th example, and $u(s)$ is the prediction function defined as
\begin{equation}\label{udef}
u(s_k) = \begin{cases}
y_+ & \text{if } s_k > s^*, \quad \quad s^* = \arg\min_{s} \{g(s)\} \\
y_- & \text{otherwise},
\end{cases}
\end{equation}
where $y_+$ and $y_-$ denote the positive and negative class labels.

Simultaneously, but independently, \citet{auacc} introduce the same metric, naming it the \textit{area under the accuracy-coverage curve} (AUACC). Coverage is defined as the percentage of performed guesses, which is exactly the definition of decision rate. In turn, AUACC is defined in its differential and weighted version, yielding
\begin{equation}    
\text{AUACC} =\int_{0}^1 \acc(j) \; d\,\cov(j), 
\label{AUACC}
\end{equation}
\noindent where 
\begin{equation}
\label{eq:acc_cov_j}
\acc(j) = \frac{\sum_{i=1}^{j} \mathbb{I}\left[y_{\pi(i)} = u(s_{\pi(i)})\right]}{\sum_{i=1}^{j} w_{\pi(i)}}, \quad
\cov(j) = \frac{\sum_{i=1}^{j} w_{\pi(i)}}{\sum_{i=1}^{N} w_{\pi(i)}},
\end{equation}
with $(s_i, y_i, w_i)$ denoting the predicted score, true label, and corresponding weight of each example, and $u(s)$, $g(s)$, and  $\pi$ taking the same definitions as in Equation \eqref{AUDRC}. The definition in Equation \eqref{AUACC} is a simplified version of that of \citet{auacc}, by assuming a fixed evaluation set and models that do not require training. Furthermore, it is clear that AUDRC can be extended to allow for weighted examples, yielding AUACC. Thus, without loss of generality, as both metrics are in fact the same, we will always refer to it as AUDRC.

Finally, \citet{tuebingen} argue that the interpretability of accuracy-decision rate curves is limited, as they disproportionately emphasize high-confidence predictions. In subsection \ref{auacc-text}, we show that AUDRC is the area under such a curve, thus inheriting this limitation.

\subsection{Invariance Properties of Evaluation Metrics}

Selecting an appropriate metric is critical for a reliable assessment of model performance. Depending on the binary classification problem at hand, factors such as dataset size and class imbalance can cause certain metrics to misrepresent a model's performance \citep{metricchoice}. In parallel, advances in deep learning loss functions seek to exploit model invariances to improve training \citep{invariancemodels}. Additionally, the nature of the input data constitutes another important factor: inputs may be subject to theoretical constraints that a metric can, and should, leverage. A clear example arises in image classification, where data augmentation via rotations is used to enhance evaluation precision \citep{imageaugmentation}.

To effectively exploit the structural properties inherent to different problems, it is essential to understand the symmetries and invariances present in performance metrics \citep{symmetriespaper}. \citet{binaryclassificationdef} identify nine distinct types of invariances, each defined in terms of transformations of the confusion matrix; the most relevant to this work is invariance $I_1$ (following the notation of those authors), which states that a measure $f([\text{TP}, \text{FP}, \text{TN}, \text{FN}])$ satisfies $I_1$ if
\begin{equation}\label{invariance1}
    f([\text{TP}, \text{FP}, \text{TN}, \text{FN}]) = f([\text{TN}, \text{FN}, \text{TP}, \text{FP}]),
\end{equation}
that is, if it is invariant under the exchange of positive and negative classes.

In contrast, \citet{metricproperties} introduce four metric properties designed to analyze the effects of class imbalance. These properties are formulated in terms of the gradients with respect to the elements of the confusion matrix—that is, they consider not only large changes in the entries of the matrix but also the sensitivity of a metric to infinitesimal perturbations thereof. For instance, what \citet{metricproperties} calls \textit{Property II} asserts that a metric is class sensitive if
\begin{equation}\label{property2}
    \frac{\partial f}{\partial \, \text{TP}} \ne \frac{\partial f}{\partial \, \text{TN}}
\quad \text{or} \quad
\frac{\partial  f}{\partial \, \text{FP}} \ne \frac{\partial f }{\partial \, \text{FN}}.
\end{equation}
In essence, this property captures whether a metric responds differently to changes in the performance of both classes when their sizes differ, rather than treating all correct and incorrect predictions uniformly.

Both \citet{binaryclassificationdef} and \citet{metricproperties} propose invariance properties that apply exclusively to categorical metrics, as they rely on explicit operations over the confusion matrix. In contrast, rank-based metrics operate on continuous scores, which are far more informative; consequently, the invariance equations become undefined for this class of binary classification metrics. 

\subsection{Contributions and Overview}

In this work, building upon invariance $I_1$ of \citet{binaryclassificationdef} and \textit{Property II} proposed by \citet{metricproperties}, we introduce the notions of \textit{local exchange of classes} (LxC) and LxC-invariant metrics. These notions are then extended to rank-based evaluation metrics. We also show that there are several machine learning problems where LxC-invariant metrics can yield a more faithful and robust performance assessment than other metrics. 

Drawing from this notion, a novel binary classification performance assessment metric, termed LxCIM (LxC-invariant metric) is introduced, which is tailored for such classification problems. Additionally, LxCIM is rank-based, thus able to capture the expressiveness of predicted scores rather than merely their correctness, while employing a fixed threshold to preserve intrinsic interpretability. Furthermore, its plot facilitates an interpretation of how performance evolves with decision rate, all while remaining computationally tractable. Finally, LxCIM has direct relationships with widely used evaluation metrics (such as AUROC, accuracy, and AUDRC), which will be presented alongside its advantages for LxC-invariant problems. Finally, the application of LxCIM to the problem of bivariate causal discovery \citep{tuebingenresults} is presented, as this problem exhibits LxC invariance, thus requiring an LxC-invariant metric. 

\section{Invariance to Local Exchange of Classes (LxC)}

\subsection{Definition for categorical metrics}\label{invariance_categorical}

For categorical metrics, which are fully characterized by the confusion matrix, invariance to local exchange of classes can be defined as
\begin{equation} \label{invariance_eq}
    f \bigl( [\text{TP},  \text{FP}, 
        \text{FN}, \text{TN} ] \bigr)
    = f \bigl(
    [ \text{TP} + \delta_1, \, \text{FP} - \delta_2, \, \text{FN} + \delta_2, \,  \text{TN} - \delta_1 ]   \bigr),
    \quad \forall\, \delta_1, \delta_2 \in \mathbb{R}.
\end{equation}
The rationale is that the metric remains unaffected by local perturbations in the assignment of positive and negative labels, highlighting its robustness to such class exchanges. Notice that the total number of correct decisions, $C = TP + \delta_1 + TN - \delta_1 = TP + TN$, for any $\delta_1$, as does the number of incorrect decisions, $I = FP + \delta_2 + FN - \delta_2 = FP + FN$, for any $\delta_2$. Notably, invariance $I_1$ proposed by \citet{binaryclassificationdef} (see Equation \ref{invariance1}) corresponds to the special case of Equation \ref{invariance_eq} for $\delta_1 = \text{TP}-\text{TN}$ and $\delta_2 = \text{FP}-\text{FN}$. Consequently, the invariance in Equation \ref{invariance_eq} is strictly stronger: whereas \citet{binaryclassificationdef} consider whether merely swapping the definitions of positive and negative classes affects the metric, LxC-invariance asserts full interchangeability of the two classes. Consequently, any metric satisfying this stronger invariance treats correct predictions in the positive and negative classes symmetrically, regardless of class balance or the proportion of correct predictions in each class.

Additionally, in differential form, this invariance can also be written as
\begin{equation}
    \frac{\partial f}{\partial \text{TP}} \neq \frac{\partial f}{\partial \text{TN}},
    \quad \text{and} \quad
    \frac{\partial f}{\partial \text{FP}} \neq \frac{\partial f}{\partial \text{FN}},
\end{equation}
making it apparent that the proposed invariance is the converse of Property II proposed by \citet{metricproperties} (see Equation \ref{property2}), referred to as class-sensitivity. LxC-invariance can thus also be described as class-insensitivity, emphasizing that LxC-invariant metrics depend solely on the relationship between the true class and the predicted score.

Moreover, any LxC-invariance metric can be expressed as a function of the total number of correct and incorrect predictions, irrespective of their class assignment. This is both a necessary and sufficient condition for LxC-invariance, as stated in the following proposition (proved in Appendix \ref{equivalence_proof}):
\begin{proposition} A function $f:\mathbb{R}^4 \rightarrow \mathbb{R}$ is an LxC-invariant metric if and only if there is some function $g:\mathbb{R}^2 \rightarrow \mathbb{R}$ such that
    \begin{equation}
    f \bigl( [
        \text{TP} ,\,  \text{FP},\, 
        \text{FN}, \, \text{TN}]
    \bigr)
    = g\bigl( [\text{TP} + \text{TN},\; \text{FP} + \text{FN}] \bigr).\label{eq:lxc}
\end{equation} \label{prop:lxc}
\end{proposition}

The only categorical metric that satisfies this invariance is accuracy. Indeed, accuracy can be expressed as the ratio of correct ($C$) decision relative to all decisions $(C+I)$:
\begin{equation}
    \text{ACC} = C/(C+I), \quad \text{where } C = \text{TP} + \text{TN}, \; I = \text{FP} + \text{FN}.
\end{equation}
Interestingly, as the total number of samples ($N$) is fixed, i.e., $\text{TP} + \text{TN} + \text{FP} + \text{FN} = C + I = N$, any other metric computed solely from $(C, I)$ is necessarily a bijective function of accuracy. This observation partly explains the absence of alternative categorical metrics in the literature that are LxC-invariant: any such metric would carry the same information as accuracy. Therefore, any rank-based metric that satisfies LxC-invariance is, to an extent, a rank-based analogue of accuracy.

\subsection{Rank-based version}
Since the definition of LxC-invariance inherently relies on the definition of correct and wrong predictions, rank-based metrics relying on scores whose meaning lies solely in their relative ordering have no possible analogous invariance. In other words, once the classical notions of TP, TN, FP, and FN are considered threshold-dependent, the invariance property cannot be defined. 

Therefore, to extend LxC-invariance to rank-based metrics, it is necessary to draw upon the concept of a decision function, as proposed for evaluation metrics that account for the reject option. The score function is paired with a bi-monotonic decision function $g(s)$ that assigns a confidence level to each prediction; as shown in Proposition \ref{prop:bimono}, this function has a global minimum at some score $s^*$. Notice also that, being strictly decreasing for all $s < s^*$ and strictly increasing for all $s > s^*$, its (set-valued) inverse satisfies: $g^{-1}(g(s^*)) = \{s^*\}$ and, for any $s\neq s^*$, $g^{-1}(g(s)) = \{s,s'\}$, where $s'\neq s$. Neither the score function nor the decision function are required to have a probabilistic interpretation. 

Consider now a function $A(s,y,w)$ that has a score, a true label, and a weight as arguments, and is defined as
\begin{equation}\label{A_eq}
A(s, y, w) =
\begin{cases}
(s^*, y, w), & \text{if } s = s^*,\\
\big(g^{-1}(g(s)) \setminus \{s\},\; 1-y, w\big), & \text{otherwise}.
\end{cases}
\end{equation}

Recall that $(s_i, y_i, w_i)$ denotes the predicted score, true label, and corresponding weight of sample $i$. The notion of rank-based LxC-invariance for some metric $f$ can be defined as follows: for any $\mathcal{I} \subseteq \{1, \dots, N\},$
\begin{equation} \label{rankinvariance} 
f \Bigl( \bigl\{ A(s_i, y_i, w_i), \; i \in \mathcal{I} \bigr\} \;\cup\; \{(s_j, y_j, w_j)\}_{j \notin \mathcal{I}} \Bigr)
= f \Bigl( \{(s_i, y_i, w_i)\}_{i=1}^N \Bigr),
\end{equation}
where the function \(A\) is defined as

In order to better understand Equation \eqref{rankinvariance}, let's start by breaking down Equation \eqref{A_eq} and the meaning of the function $A$. In large, the objective of the function $A$ is to perform the exchange of classes for the tuple $(s, y, w)$. In categorical metrics, exchanging classes simply requires changing simultaneously the predicted and true labels. In the rank-based domain, however, the score is also relevant as a measure of the method's confidence in its prediction (given by the function $g(s)$). Consequently, performing the exchange of classes boils down to simultaneously switching the true label ($y'=1-y$) and the score to a score that induces an equivalent level of confidence in the other class. This is done by noticing that, given the ordinal meaning at the basis of rank-based metrics, and the necessary bi-monotonicity of the decision function (see Proposition \ref{prop:bimono}), \(g^{-1}(g(s)) = \{s, s'\}\) for all \(s \in \mathbb{R} \setminus \{s^*\}\). Essentially, for all scores with the exception of the decision threshold $s^*$, the only other score value that shares the same level of confidence in its prediction, is, necessarily, associated with the other class. At the decision threshold, the prediction has null confidence, and therefore, its predicted class is undefined; thus changing or not the true label carries no meaning. Therefore, in order to perform the exchange of classes, $A$ simultaneously flips the true label, while assigning the only other score with the same confidence level (thus belonging to the other class) as the score. Naturally, the weights are kept the same. In turn, Equation \eqref{rankinvariance} introduces the concept of local to the exchange of classes defined by $A$, by stating that the equality should hold regardless of the examples that are exchanged.

In all, Equation~\eqref{rankinvariance} states that a metric is LxC-invariant if and only if, for any subset of samples, a simultaneous inversion of the labels, with a corresponding adjustment of its scores such that the new score reflects the same level of confidence in the opposite class as the original score did in its class (and vice versa), the metric's value remains unchanged. Put differently, a metric is invariant to local exchange of classes if it depends solely on how well the ordered predictions in terms of the confidence induced by the scoring and decision functions align with the true labels. This captures the rank-based generalisation of LxC-invariance.

Regardless of the chosen decision function, AUROC does not generally satisfy the invariance property defined in Equation~\eqref{rankinvariance}. Since AUROC measures $\mathbb{P}[s(X^+) > s(X^-)]$, in order to remain unchanged under the simultaneous label and score transformation induced by $A(\cdot)$ in Equation~\eqref{A_eq}, the transformed score would need to occupy the symmetric position to its original value within the ordered scores, a condition that is generally not satisfied. A simple counterexample illustrates this point: consider $g(s) = |s|$, $s = [-4, -3, 1, 2]$, and $y = [0,1,0,1]$. By selecting the subgroup containing only $(s_3, y_3)$ and applying function $A(\cdot)$, an AUROC of $1$ is obtained, since $s' = [-4, 1, 2, 3]$ and $y' = [0,0,1,1]$.

A more extreme scenario arises when the selected subgroup consists entirely of negative examples. In this case, AUROC becomes undefined, as its computation requires the existence of both true labels (for an infintely low threshold, all samples would be classified as false negatives, leading to a division by zero). More generally, any rank-based metric that is sensitive to class imbalance, such as the KS statistic or AUPRC, cannot satisfy the criterion specified in Equation~\eqref{rankinvariance}.

\subsection{LxC-invariant problems}\label{invariant_examples}

Having defined LxC-invariance, several characteristics can help to motivate when such invariance is appropriate; e.g., if class proportions are irrelevant, if classes are interchangeable, or if accuracy is the best categorical metric for quantifying a method's performance. More generally, this intuition can be formalized into two well-defined scenarios in which an LxC-invariant metric is appropriate:
\begin{itemize}\label{invariance_guidelines}
    \item \textbf{Models are encouraged to learn class-specific biases:} If the classes themselves carry no intrinsic meaning, it is irrelevant whether one class is classified more accurately than another. In such cases, the goal is only to evaluate overall performance across all classes, in accordance with their distribution in the evaluation set. Metrics used in this scenario should be LxC-invariant to ensure that they measure the method's performance exclusively, without being influenced by class proportions or any predictive asymmetries present between classes.
    \item \textbf{Models are prevented from learning class-specific biases:} Alternatively, if a model is, by prior knowledge, unable to learn biases toward specific classes, any asymmetries in evaluation between them are inherently random. Accounting for these asymmetries would amount to measuring the randomness of the evaluation task rather than the method's true performance. Consequently, LxC-invariant metrics are also appropriate in this setting, since they explore the model's structure to ensure they focus evaluation solely on its performance.
\end{itemize}

A class of machine learning problems that requires LxC-invariant metrics is pairwise choice tasks. In these problems, the goal is to select between two alternative variables (e.g., predicting which of two football teams will win a game). More generally, any problem in which a simple and widely accepted transformation of the input flips the label can be viewed as a pairwise choice problem. These tasks should be evaluated using metrics that are LxC-invariant, since each sample can be arbitrarily labelled depending solely on the input ordering, leading to the first criterion outlined in the proposed guidelines being fulfilled (see subsection~\ref{invariance_guidelines}). Moreover, due to this inherent arbitrariness\footnote{Pairwise choice evaluation datasets often come with a single true class (e.g., all match winners appear first).}, the labelling process itself becomes a part of the evaluation. Because, in this context, labelling introduces randomness, the use of non-LxC invariant metrics renders the evaluation process non-deterministic. That is, repeated evaluations on the same test set may yield slightly different outcomes.

For instance, in bivariate causal discovery\footnote{This field will be the canonical example discussed in Section~\ref{causalexample}.} \citep{tuebingen}, the problem can be framed as choosing which of two statistical variables is more likely to be the cause (or, equivalently, the effect). Another example of a pairwise choice problem arises in sports prediction, where the objective is to determine which of two players is more likely to win a match \citep{tennisexample}.

A different class of machine learning problems suitable to evaluation using LxC-invariant metrics is item recommendation. In it, the goal is to rank items according to user preference \citep{itemrecommendation}. One known approach is to employ search algorithms on top of a pairwise choice model that, given two items, predicts which one should be ranked higher. This reduces to a binary classification problem, with applications spanning from ranking internet search results \citep{searchresultpairsranking} to assessing image quality \citep{imagequality}.

On the other hand, an example corresponding to the second criterion of the proposed guidelines (see subsection~\ref{invariance_guidelines}) is the evaluation of NAND memory performance, based on detected bit flips \citep{nandbitprediction}. NAND memories can be modelled as a binary symmetric channel (BSC) \citep{shannon}, implying that a $0\to1$ flip is equally as likely as a $1\to0$ flip. Consequently, any observed asymmetry in bit flips during evaluation is known to be inherently random and should not affect the evaluation of each memory.

\section{LxCIM}

\subsection{Definition and Interpretation}

\begin{figure}[h!]
    \centering
    \includegraphics[width=0.45\linewidth]{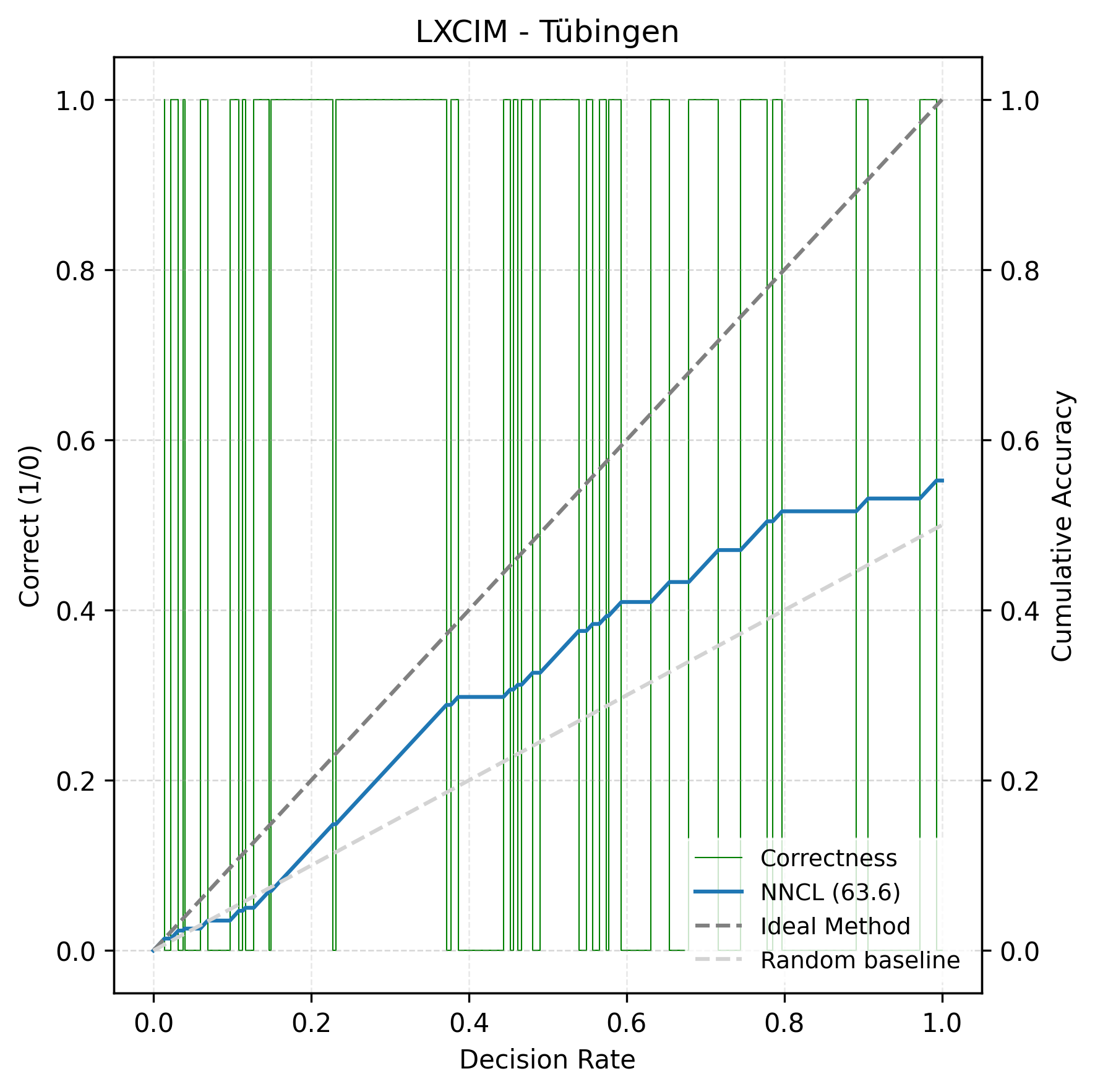}
    \includegraphics[width=0.45\linewidth]{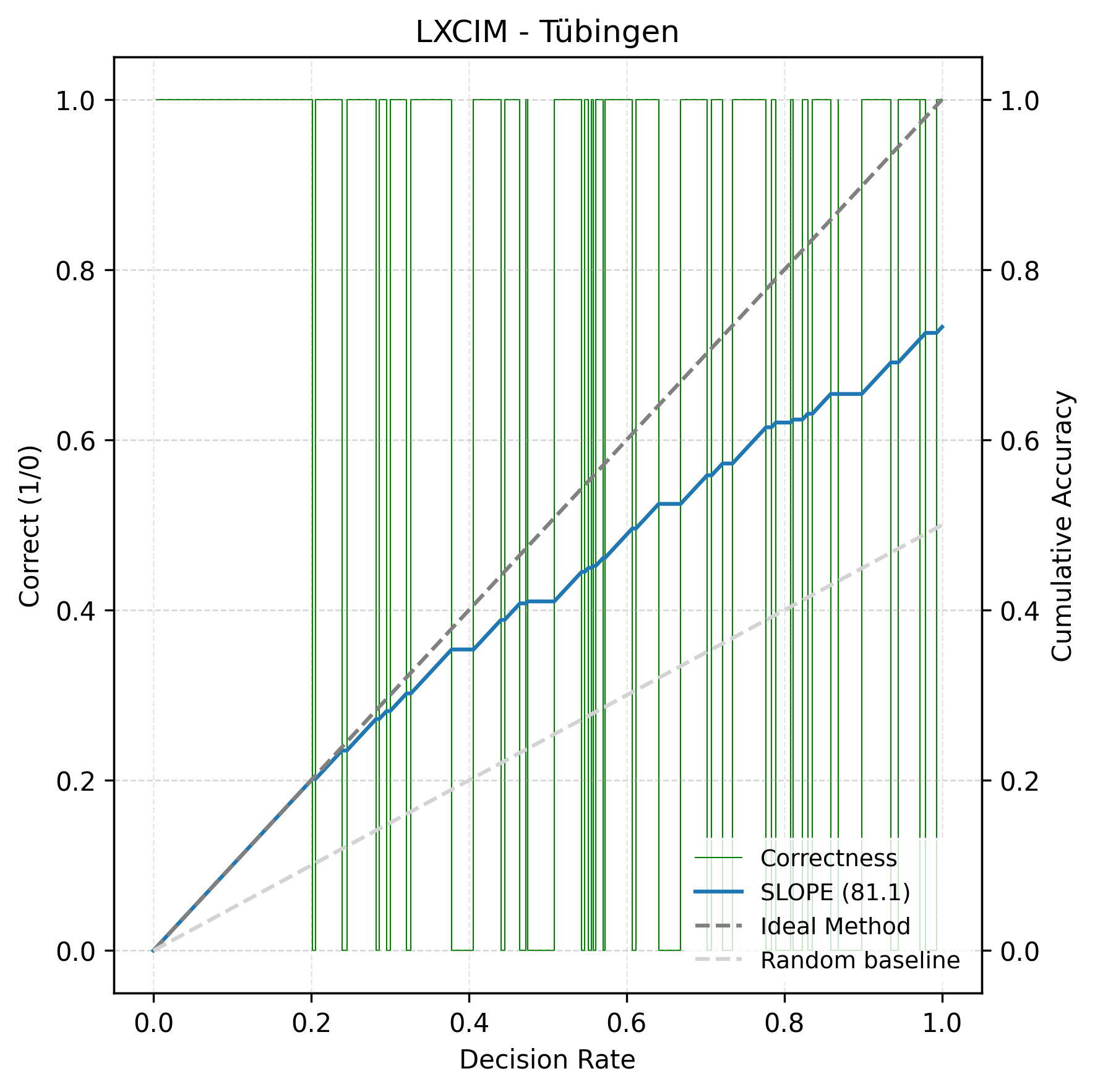}
    \caption{The curves for both $G(w)$ and $\mathbb{I} \left[ u(s_{j(w)}) = y_{j(w)} \right]$ as a function of the weighted ordered decision rate are plotted for the evaluation of both NNCL \citep{nncl}, on the left, and SLOPE \citep{slope}, on the right, on the T\"ubingen dataset \citep{tuebingen} (a bivariate causal discovery problem). In turn, LxCIM corresponds to the area under the $G(w)$ curve.}
    \label{alameda_mock}
\end{figure}

LxCIM is defined as twice the \textit{area under the cumulative accuracy-decision rate curve}. Alternatively, it can be measured as twice the cumulative sum of the weighted accuracy computed over increasing confidence levels:
\begin{equation}\label{alameda_discrete}
\text{LxCIM} = \frac{2}{\sum_{i=1}^N w_i} \sum_{i=1}^N \sum_{j=1}^i w_{\pi(j)} , \mathbb{I}\big[y_{\pi(j)} = u(s_{\pi(j)})\big].
\end{equation}
where $(s_i, y_i, w_i)$ denotes the predicted score, true label, and corresponding weight of each example, and $u(s),g(s), \text{and } \pi$ take the same definitions of Equation \eqref{AUDRC}.  By expanding the inner sum in Equation~\eqref{alameda_discrete}, it can be rewritten as:
\begin{equation}\label{alameda_discrete2}
    \text{LxCIM} = \frac{2}{\sum_{i=1}^N w_i} \sum_{i=1}^N (N - i) \, w_{\pi(i)} \, \mathbb{I}\big[y_{\pi(i)} = u(\mathbf{s}_{\pi(i)})\big].
\end{equation}
In its differentiable form, LxCIM can be expressed as:
\begin{equation}\label{alameda_eq}
    \text{LxCIM} = \frac{2}{\sum_k w_k} \int_{0}^{\sum_k w_k} G(i) \, di, 
\end{equation}
where $G(i)$ is the cumulative accuracy as a function of decision rate, 
\begin{equation}
    G(i) = \int_{0}^{i} \mathbb{I} \big[ u(s_{j(w)}) = y_{j(w)} \big] \, dw, 
\end{equation}
and $j(w)$ computes the index of the tuple $(s_j,y_j,w_j)$ under evaluation for the weighted decision rate $w$,
\begin{equation}
    j(w) = \arg\max_{k} \left\{ \sum_{l=1}^k w_{\pi(l)} < w \right\}.
\end{equation}
LxCIM admits a natural visual representation as twice the area under the $G(i)$ curve, appropriately named the cumulative accuracy-decision rate curve, as illustrated in Figure~\ref{alameda_mock}. 

In Equation \eqref{alameda_discrete}, LxCIM can be interpreted as first ordering all scores according to their confidence as determined by the decision function $g(s)$. Then, for each level of confidence, it computes the cumulative accuracy up to that point of the predicted scores. This is equivalent to computing the accuracy under the assumption that all predictions taken with lower confidence than that considered are incorrect. Lastly, the area under this cumulative accuracy-decision rate curve is half of LxCIM. Moreover, the initial multiplication by $2$ serves as a normalization factor, ensuring that a random predictor achieves a value of $0.5$ while a perfect predictor attains $1$. This intuition is visually reinforced in Figure~\ref{alameda_mock}, where the area under the curve for a perfect method is exactly $0.5$. Furthermore, note that $G\big(\sum_k w_k\big) = \text{ACC}$: when all examples are included, the cumulative accuracy naturally coincides with the overall accuracy of the method.

In turn, Equation \eqref{alameda_discrete2} makes explicit that each correctly classified instance contributes to the overall metric in proportion to the number of samples that follow it in the ranked sequence. From both expressions, it becomes evident that the relative importance assigned to each prediction, given by the term $w_{\pi(i)} \cdot \mathbb{I}\big[y_{\pi(i)} = u(\mathbf{s}_{\pi(i)})\big]$, scales linearly with $(1 - i/N)$ where $i$ is the respective decision rate. This linear weighting with respect to rank embodies LxCIM’s inherently rank-based design, ensuring that the metric not only accounts for correctness but also linearly integrates the relative confidence structure of the model’s outputs.

Furthermore, the $G(w)$ curve provides additional interpretive power. An ideal method contains a constant slope of $1$, while a random baseline exhibits a slope of $0.5$. Consequently, the slope at any point on the curve conveys the expected accuracy for predictions performed at the corresponding confidence level. This enables interpretations analogous to reject-option metrics, where a threshold on the decision function is used to abstain from low-confidence predictions. The LxCIM plot makes this threshold analysis visually transparent: regions with higher slopes indicate reliable predictions, while flatter regions highlight predictions approaching randomness. For instance, in Figure \ref{alameda_mock}, while NNCL approaches a slope nearing $0.5$, corresponding to random guessing, around the $60\%$ decision rate, SLOPE displays a slope significantly above $0.5$ across the entire decision rate. Consequently, it is easy to analyse at which decision rate a method stops making meaningful predictions. This fine-grained, rank-based interpretability of model performance as a function of decision rate represents one of LxCIM’s principal advantages.



Finally, it is necessary to prove that the metric is LxC-invariant. This means, for any subset \(I \subseteq \{1, \dots, N\}\):
\begin{equation}
\begin{aligned}
\text{LxCIM}\!\,\Big(\{(A(s_i,y_i,w_i))\}_{i \in I} \;\cup\; \{(s_j,y_j,w_j)\}_{j \notin I}\Big)
&= \text{LxCIM}\!\,\Big( \{(s_i, y_i, w_i)\}_{i=1}^N\Big) \\
\iff 
G\Big(w_k, \{(A(s_i,y_i,w_i))\}_{i \in I} \;\cup\; \{(s_j,y_j,w_j)\}_{j \notin I}\Big)
&= G\Big(w_k, \{(s_i, y_i, w_i)\}_{i=1}^N\Big),
\end{aligned}
\end{equation}
since $\text{LxCIM} = \frac{2}{\sum_k w_k} \int_{0}^{\sum_k w_k} G(i) \, di$, and only $G(i)$ depends on $\{(s_j,y_j,w_j)\}$. 

Therefore, for this equality to hold, $G(i)$ must be invariant under the function $A(\cdot)$ (defined in Equation \eqref{A_eq}). Recall that $G(i) = \int_{0}^{i} \mathbb{I} \big[ u(s_{j(w)}) = y_{j(w)} \big] \, dw$. Knowing the weights remain unchanged under $A(\cdot)$, it suffices to show that the permutation $\pi$ remains identical and that the indicator function $\mathbb{I} \big[ u(s_{j(w)}) = y_{j(w)} \big]$ is preserved. 

Regarding the indicator function, let $s'$ denote the transformed score after applying $A(\cdot)$. Since $g(s) = g(s')$, $u(s') = 1 - u(s)$ under the standard notation $\{y_+, y_-\} = \{1,0\}$. Consequently,
\begin{equation}
    \mathbb{I} \big[ u(s_{j(w)}') = y_{j(w)}' \big] = \mathbb{I} \big[ 1-u(s_{j(w)}) = 1-y_{j(w)} \big] = \mathbb{I} \big[ u(s_{j(w)}) = y_{j(w)} \big].
\end{equation}
Additionally, the permutation $\pi$ is unaffected by $A(\cdot)$, since $g(s') = g\big(g^{-1}(g(s)) \setminus \{s\}\big) = g(s)$ by definition.  
Hence, it follows directly that LxCIM is LxC-invariant as defined in Equation~\eqref{invariance_eq}.

Finally, as discussed in subsection~\ref{invariance_categorical}, accuracy is the only categorical metric that satisfies LxC-invariance. Even though rank-based approaches allow for greater flexibility and multiple different rank-based metrics can achieve LxC-invariance (such as AUDRC), this similarity underscores the observation that LxCIM is most appropriately applied to the same problems where accuracy is the ideal categorical metric. In this regard, LxCIM can be interpreted as a natural rank-based analogue of accuracy, preserving the same invariance property while leveraging the additional information encoded in predicted scores.

\subsection{Relationship with other metrics}

\subsubsection{AUROC}\label{auroc_relantionship}

LxCIM exhibits two clear differences from AUROC. Firstly, its invariance to local exchange of classes (LxC) provides a significant benefit in tasks that satisfy either of the two criteria outlined in subsection~\ref{invariance_guidelines}.

Secondly, LxCIM provides a clear and interpretable representation of how expected performance evolves with the decision rate. In contrast, the ROC curve supports no such interpretation, as no fixed decision threshold is considered. In other words, the concepts of confidence or predicted class are not well-defined. Even if one were to choose a specific decision threshold, the ROC curve would still not provide an easy interpretation, since positive and negative predictions are plotted separately across the $\text{TPR} = 1 - \text{FPR}$ line. Such a representation can obscure biases in scoring one class with higher than the other, making it more challenging to assess the expected performance under a reject-option scenario.


Nevertheless, there are many significant similarities between both metrics, which are best identified by noting that LxCIM is mathematically equivalent to the AUROC computed over a dataset $\Omega$ where every original sample is duplicated, with each duplicate having the opposite label, a complementary score, and the same confidence level. This duplication process can be expressed using the function $A(\cdot)$ (see Equation~\ref{A_eq}):
\begin{equation}\label{eq:Omega}
\Omega = \{(s_i, y_i, w_i)\}_{i=1}^N \cup \{A(s_i, y_i, w_i)\}_{i=1}^N.
\end{equation}
Consequently,
\begin{equation}\label{doubling_eq}
    \mathrm{LxCIM}\!\left( \{(s_i, y_i, w_i)\}_{i=1}^N \right)
    = \mathrm{AUROC}\!\left(\Omega \right).
\end{equation}

To better understand Equation~\eqref{doubling_eq}, consider the simplest case\footnote{Since there exists a bijective mapping between this function and any general decision function $g$, this analysis does not entail any loss of generality.}, where $g(s) = |s|$. Let $\bar\Omega$ denote the $2N$-tuple 
\begin{equation}
    \bar\Omega = \bigl( (s_{\tau(1)},y_{\tau(1)},w_{\tau(1)}), ...,  (s_{\tau(2N)},y_{\tau(2N)},w_{\tau(2N)})\bigr) \equiv \bigl( (\bar{s}_1,\bar{y}_1,\bar{w}_1), ...,  (\bar{s}_{2N},\bar{y}_{2N},\bar{w}_{2N})\bigr)
\end{equation}
where $\tau$ is the permutation of $\{1,...,2N\}$ that sorts the elements of $\Omega$ in decreasing order of the $s_i$. In turn, the score and label vectors exhibit perfect anti-symmetry about the midpoint, while the weight vector is symmetric:
\begin{equation}
    \forall i \in \{1, \dots, 2N\}, \quad \bar{s}_{N-i} = -\bar{s}_{N+i}, \quad \bar{y}_{N-i} = 1 - \bar{y}_{N+i}, \quad \bar{w}_{N-i} = \bar{w}_{N+i}.
\end{equation}

\begin{figure}[h!]
    \centering
\includegraphics[width=0.45\linewidth]{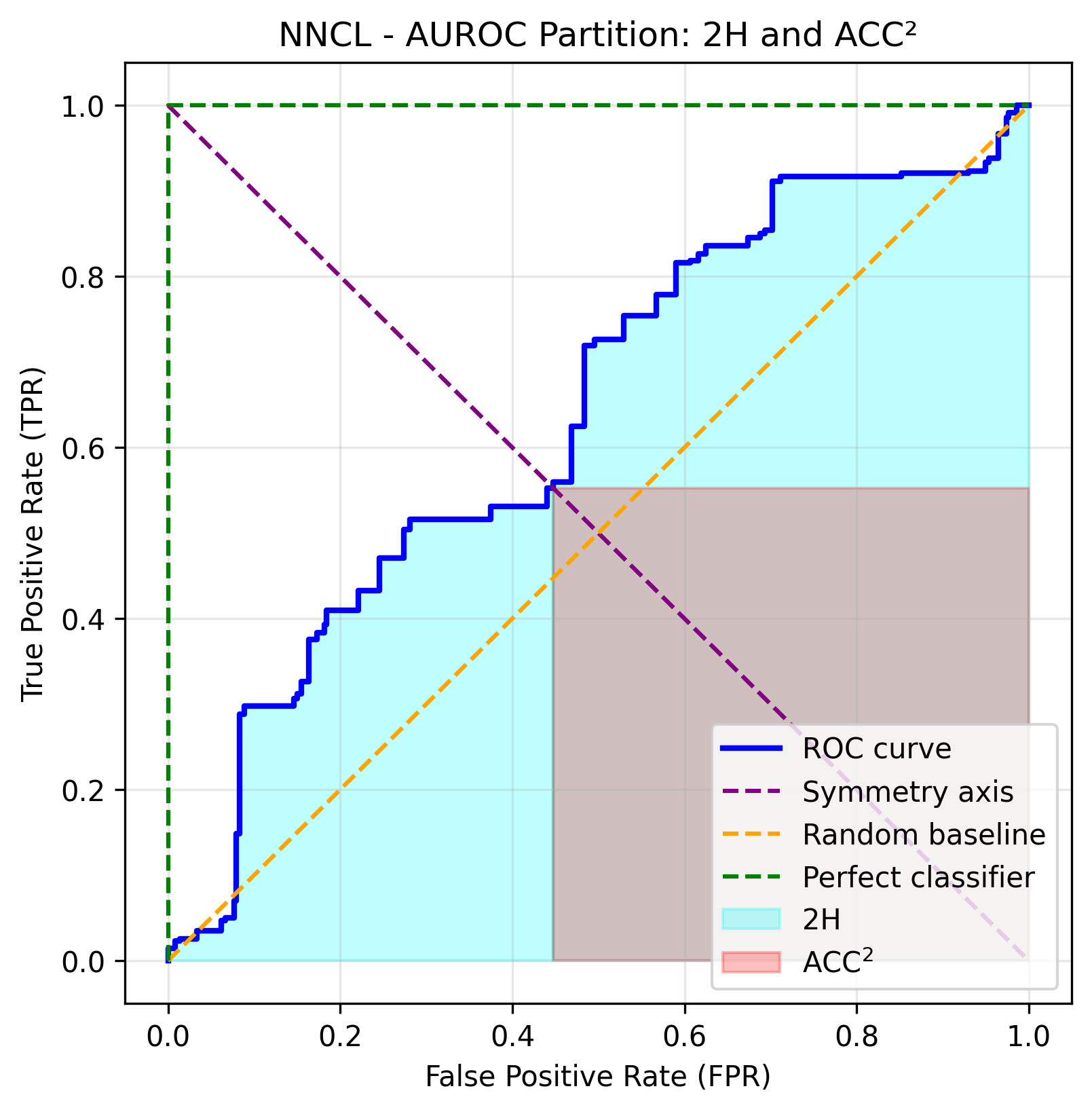}
\includegraphics[width=0.45\linewidth]{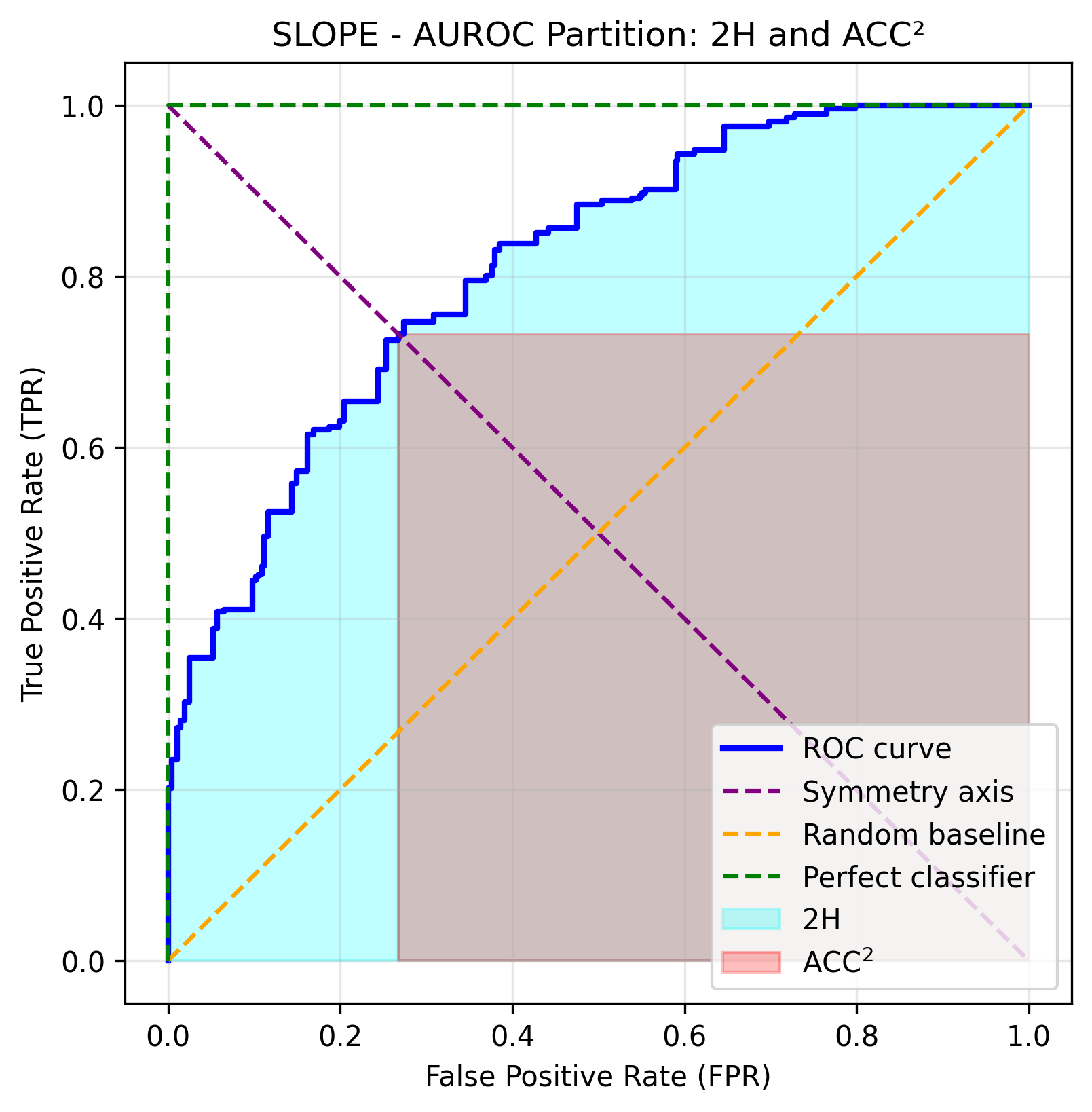}
    \caption{The ROC curves for both NNCL and SLOPE (left and right, respectively) evaluated on the augmented T\"ubingen dataset according to Equation \eqref{doubling_eq}. For context, the performance of both an ideal method and a random baseline is drawn, alongside the symmetry axis created by the duplication procedure.}
    \label{symauroc_mock}
\end{figure}

Due to the symmetry of the augmented dataset around its midpoint, the two curves $\text{TNR}(\text{FNR})$ and $\text{TPR}(\text{FPR})$ coincide exactly. This arises because traversing the ordered vector from left to right or from right to left produces identical ROC curves. Consequently, given that $\text{TPR} = 1 - \text{FNR}$ and $\text{TNR} = 1 - \text{FPR}$, the ROC curve $\text{TPR}(\text{FPR})$ must be symmetric with respect to the line $\text{TPR} = 1 - \text{FPR}$. This symmetry is shown in Figure~\ref{symauroc_mock}. Moreover, the point where AUROC intersects the line $\text{TPR} = 1 - \text{FPR} = \text{ACC}$. 

\begin{proposition} For any $\Omega$, the ROC curve intersects the line $\mathrm{TPR} = 1 - \mathrm{FPR}$ at the point
\begin{equation}
    (1 - \mathrm{ACC},\ \mathrm{ACC}).
\end{equation}
\label{eq:crossingpoint}
\label{prop:crossingpoint}
\end{proposition}
A detailed proof is provided in Appendix~\ref{symmetry_axis_proof}. Intuitively, this can be understood as corresponding to computing  $\text{TPR}$ and $\text{FPR}$ at the decision threshold $s = s^* = 0$. 

A deeper intuition can be gained by noting that sweeping the decision threshold from $+\infty$ to $0$ in the augmented dataset corresponds to traversing the entries in order of descending confidence, mirroring exactly the procedure present in LxCIM. Nevertheless, in the duplicated dataset, the labels and scores do not correspond directly to the original points when sorted by $g(s)$. Specifically, all points in the duplicated dataset satisfy $u(s_{\tau(i)}) = y_+$, for any  $i \in \{1,...,N\}$. However, a careful examination of the augmentation process (see Equations~\eqref{A_eq} and \eqref{eq:Omega}) shows for all points, $\mathbb{I} \big[ u(s_{\pi(i)}) = y_{\pi(i)} \big] = \mathbb{I} \big[ u(\bar{s}_i) = \bar{y}_i\big]$, for $i=1,...,N$. For the duplicated points, although the predicted class is flipped, the corresponding label is simultaneously inverted by $A(\cdot)$, ensuring that $\mathbb{I} \big[ u(s_{\pi(i)}) = y_{\pi(i)} \big]$ remains unchanged in the augmented dataset.
In other words, the first half of the duplicated dataset can be interpreted as the original dataset with the operation $A(\cdot)$ applied solely to the examples originally predicted as negative (and conversely for the second half of $\bar{\Omega}$). This construction represents a specific instance of Equation~\eqref{rankinvariance}, which formalizes LxC-invariance. As a result, from the perspective of LxCIM, which is by definition LxC-invariant, evaluating the original dataset is mathematically equivalent to evaluating the first half of the duplicate dataset $\bar{\Omega}$. 

This equivalence provides the crucial insight behind the equality between the two metrics stated in Equation~\eqref{doubling_eq}. More so, by naming the blue-shaded region in Figure~\ref{symauroc_mock} as $2H$, the area $H$ can be computed solely from the analysis of $\bar{\Omega}|_{i=1:N}$, leveraging the symmetry previously established. With this definition of $H$, $\text{AUROC} \left(\Omega
\right)
    = \mathrm{ACC}^2 + 2H$. The original claim present in Equation \eqref{doubling_eq} is equivalent to the following proposition (proved in Appendix~\ref{doublingproof}).
    
\begin{proposition} \label{prop:doubling}
 For any duplicated dataset, $\Omega$, $\mathrm{LxCIM}\!\left(\{(s_i, y_i, w_i)\}_{i=1}^N\right) = \mathrm{ACC}^2 + 2H.$
\end{proposition}

Consequently, LxCIM can be interpreted as an adaptation of AUROC tailored for problems that demand LxC invariance. This interpretation is justified by the fact that, for any dataset where AUROC exhibits symmetry, both AUROC and LxCIM yield identical results. Therefore, LxCIM effectively enforces this symmetry even in scenarios where it may not be present in the evaluation data, ensuring that the metric remains consistent with the LxC-invariant objective.

\subsubsection{AUDRC}\label{auacc-text}

The weighted and differential form of the \textit{area under the accuracy-decision rate curve} (AUDRC) expressed in Equation \eqref{AUDRC} can be alternatively written as
\begin{equation}\label{AUDRC_similarity}
\begin{aligned}
\text{AUDRC} &= \frac{1}{\sum_k w_k} \int_{0}^{\sum_k w_k} acc(i) \, di, \\
\text{with} \quad acc(i) &= \frac{1}{i} \int_{0}^{i} \mathbb{I} \big[ u(s_{j(w)}) = y_{j(w)} \big] \, dw = \frac{G(i)}{i},
\end{aligned}
\end{equation}
where $G(i)$ is defined as in Equation~\eqref{alameda_eq} and $\frac{d\,cov}{dj} = \frac{1}{\sum_k w_k}$. 

The single key difference bewteen LxCIM and AUDRC lies in their treatment of the cumulative information: AUDRC integrates the accuracy computed up to the considered decision rate, whereas LxCIM considers the cumulative accuracy at that point, effectively treating all remaining points with lower confidence as misclassified.

Additionally, by virtue of their structural similarity, AUDRC also inherits LxC-invariance, with the same reasoning and derivation applied to LxCIM. The presence of the normalisation term $\frac{G(i)}{i}$ does not affect this property, as it functions purely as a scaling factor that does not interact with, nor alter, the effects of any local exchange of classes.

\begin{figure}[h!]
    \centering
    \includegraphics[width=0.45\linewidth]{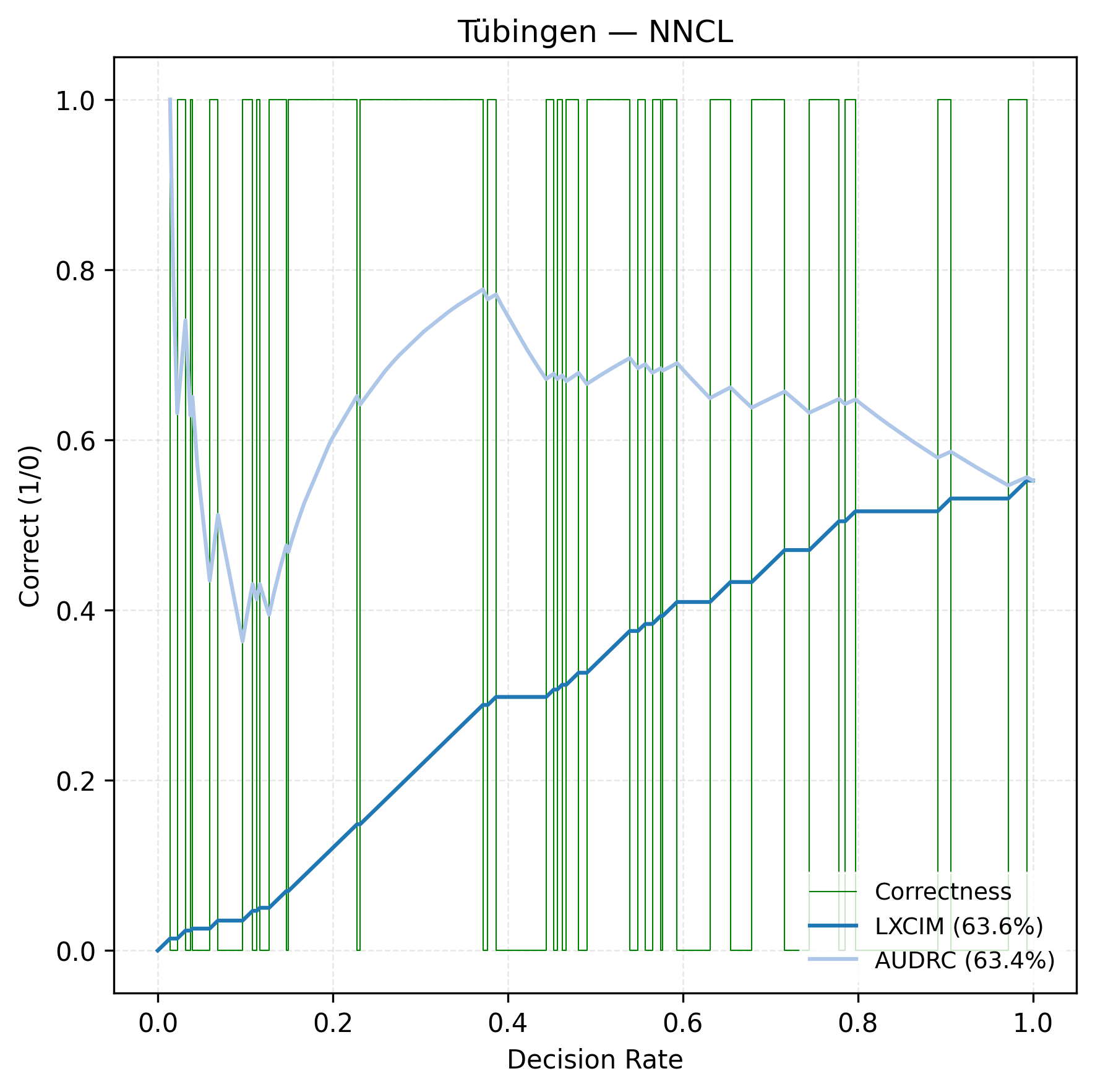}
    \includegraphics[width=0.45\linewidth]{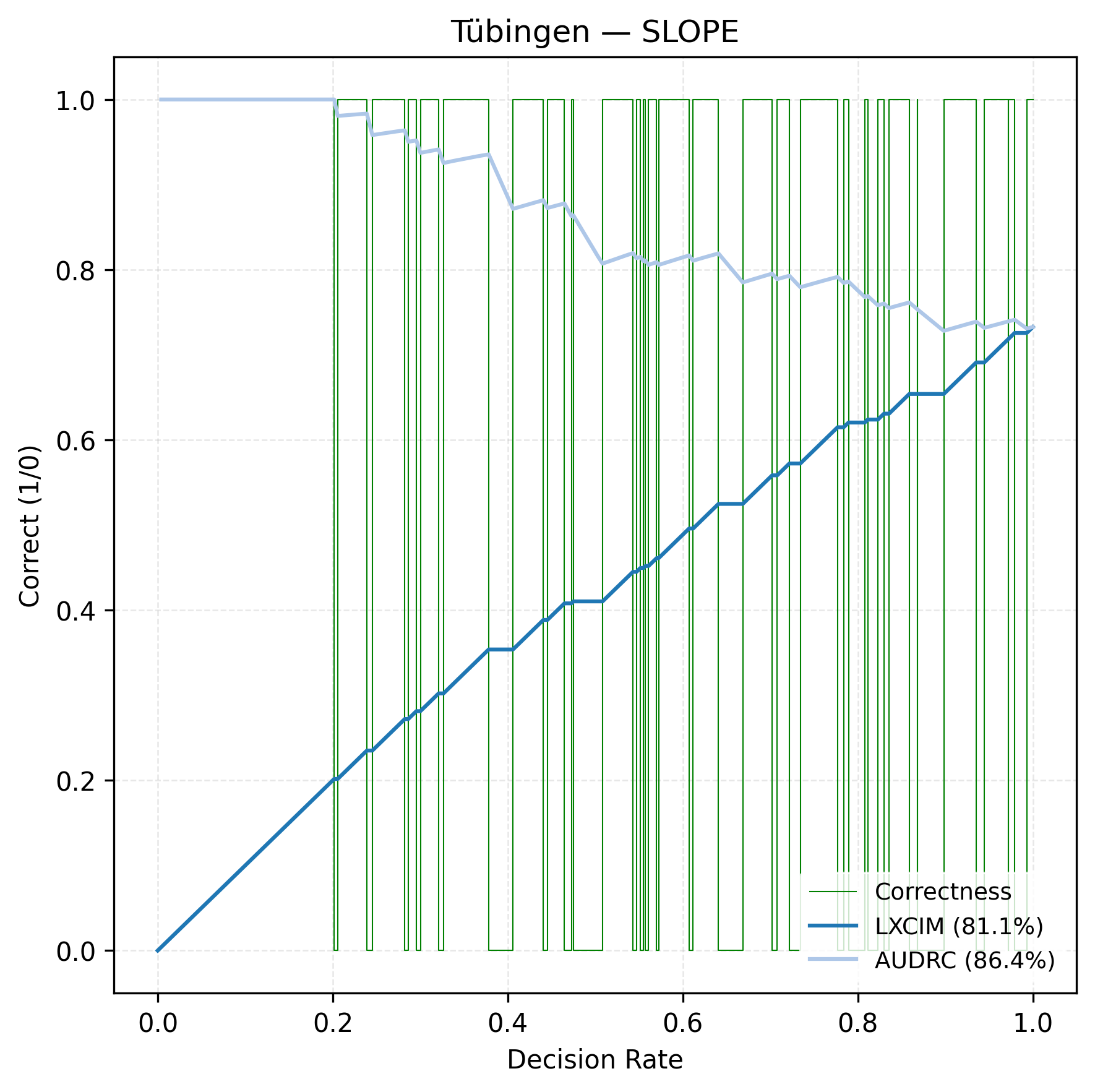}
    \caption{The curves of both AUDRC and LxCIM as a function of decision rate for both NNCL and SLOPE evaluated on the T\"ubingen dataset. Additionally, and for context, the correct and wrong guesses are also visualised as a function of the decision rate. Lastly, the performance of an ideal method and a random baseline is provided.}
    \label{auacc_mock}
\end{figure}

In turn, the accuracy-decision rate curve aims to provide a detailed view of how accuracy evolves as a function of decision rate. Figure \ref{auacc_mock} displays both the cumulative accuracy and accuracy as a function of the decision rate (LxCIM and AUDRC curves, respectively) for the same experimental setup present in Figures \ref{alameda_mock} and \ref{symauroc_mock}. 

In all, LxCIM distinguishes itself from AUDRC along three key points: (1) the richer interpretability afforded by the cumulative accuracy-decision rate curve, (2) the relative importance assigned to each prediction as a function of the considered decision rate, and (3) the guaranteed computability of LxCIM, whereas AUDRC fails in the asymptotic limit $N \to \infty$.

Firstly, when comparing the curves of both metrics displayed in Figure \ref{auacc_mock}, the interpretation of performance in terms of decision rate exhibits a subtle yet significant difference. For AUDRC, the value of the curve at a given decision rate indicates the expected accuracy if that decision rate were chosen. In contrast, LxCIM’s cumulative accuracy–decision rate curve conveys a more interesting insight: the slope at a particular decision rate directly reflects the model’s expected accuracy for predictions whose confidence lies in the $1-$``decision rate" percentile. 


Alternatively, in AUDRC, accuracy naturally decreases as decision rate increases, making the trade-off between expanded coverage and diminishing accuracy less transparent.
For instance, considering the Figure \ref{auacc_mock}, for NNCL, AUDRC displays a significantly decreasing accuracy from the $40\%$ decision rate onwards, while in fact this initial dip is a consequence of several consecutive failed guesses, but only from the $60\%$ percentile does the slope of LxCIM near $0.5$. On the other hand, since SLOPE performs all its initial guesses correctly, it seems that its performance suffers significantly from the $20\%$ decision rate onwards, despite performing significantly above random for such decision rates. This is notably visible in the LxCIM curve, where SLOPE's slope stays significantly above $0.5$ for all decision rates. Furthermore, both AUDRC curves (NNCL and SLOPE) seem to be equally steep in the second half, creating the illusion that both struggle equally with increasing coverage. However, the analysis of LxCIM makes clear that this is not the case, as SLOPE maintains consistently higher slopes than NNCL. In turn, the illusion is a sole consequence of SLOPE starting from a higher accuracy. Everything considered, LxCIM allows a more nuanced and actionable interpretation of how model performance evolves with decision rate.

On another hand, by unrolling its inner summation, to align with Equation~\eqref{alameda_discrete2} of LxCIM, the following equivalent equation for AUDRC is obtained:
\begin{equation}
    \text{AUDRC} = \frac{1}{\sum_{i=1}^N w_i} \sum_{i=1}^N \frac{(N-i)}{i} \, w_{\pi(i)} \cdot \mathbb{I}\left[y_{\pi(i)} = u(\mathbf{s}_{\pi(i)})\right].
\end{equation}

\begin{figure}[h!]
    \centering
    \includegraphics[width=\linewidth]{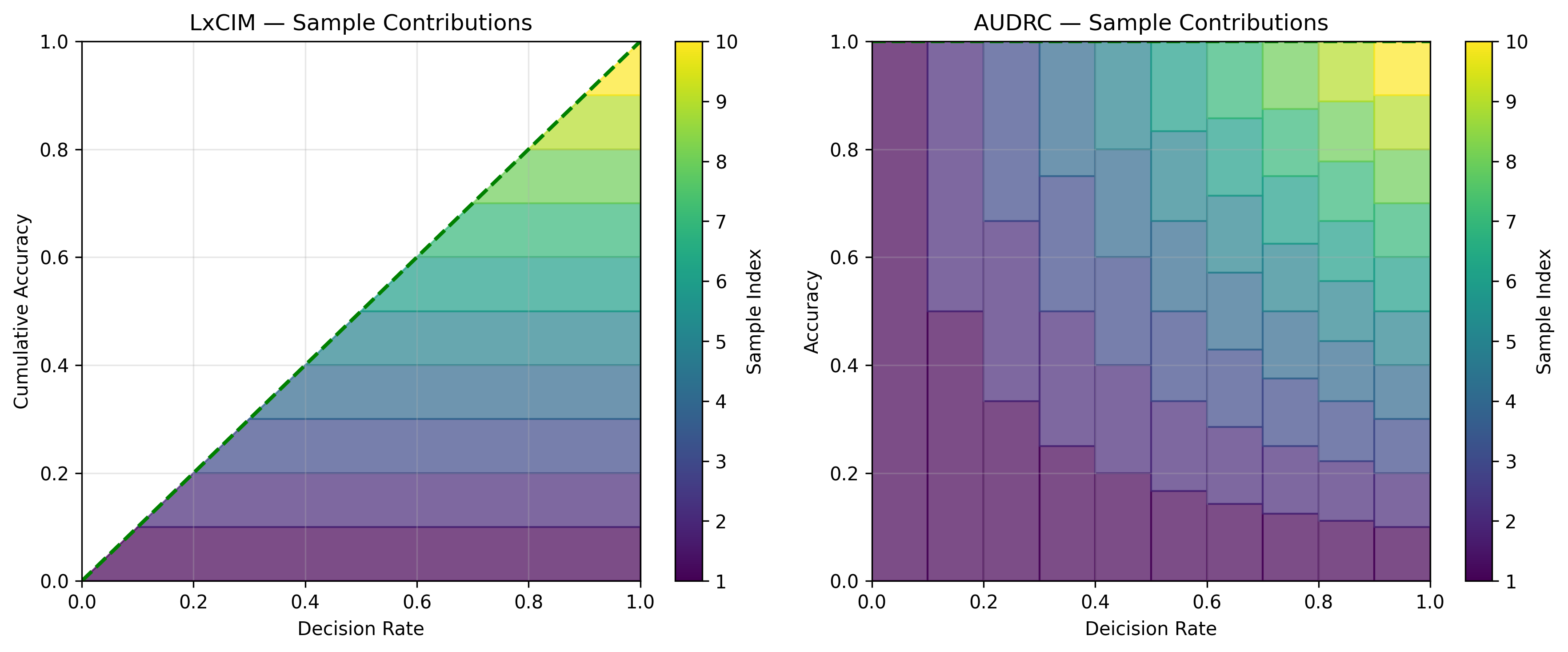}
    \caption{The relative importance of each sample for both LxCIM (on the left) and AUDRC (on the right) is plotted. In order to do so, an experiment consisting of two perfect classifiers, evaluated on a uniformly weighted dataset of 10 samples, was performed. The colors reflect the contribution of each sample to the total area under the curve.}
    \label{sample_contributions}
\end{figure}

In LxCIM, the relative importance assigned to each prediction as a function of decision rate is proportional to $N - i$. In contrast, for AUDRC, this weighting scales as $N/i$. Consequently, AUDRC places disproportionately greater emphasis on higher-confidence predictions, while diminishing the contribution of lower-confidence ones. This behaviour accentuates the limitation previously highlighted for the accuracy-decision rate curve, namely its tendency to overemphasise the value of top-ranked predictions.
In Figure \ref{sample_contributions}, the relative contribution of each correctly predicted sample to the total metric, the area under the curve, is visualized for both LxCIM and AUDRC.

Additionally, it is interesting to note that, in rank-based evaluation, the raw scores are transformed into an ordered ranking, which inherently induces a linear scaling of confidence across the examples in the test set, irrespective of the original score magnitudes. LxCIM’s linear weighting aligns naturally with this transformation, ensuring that the metric’s emphasis on each prediction is consistent with the ordinal interpretation of the scores.

\begin{figure}[h!]
    \centering
    \includegraphics[width=\linewidth]{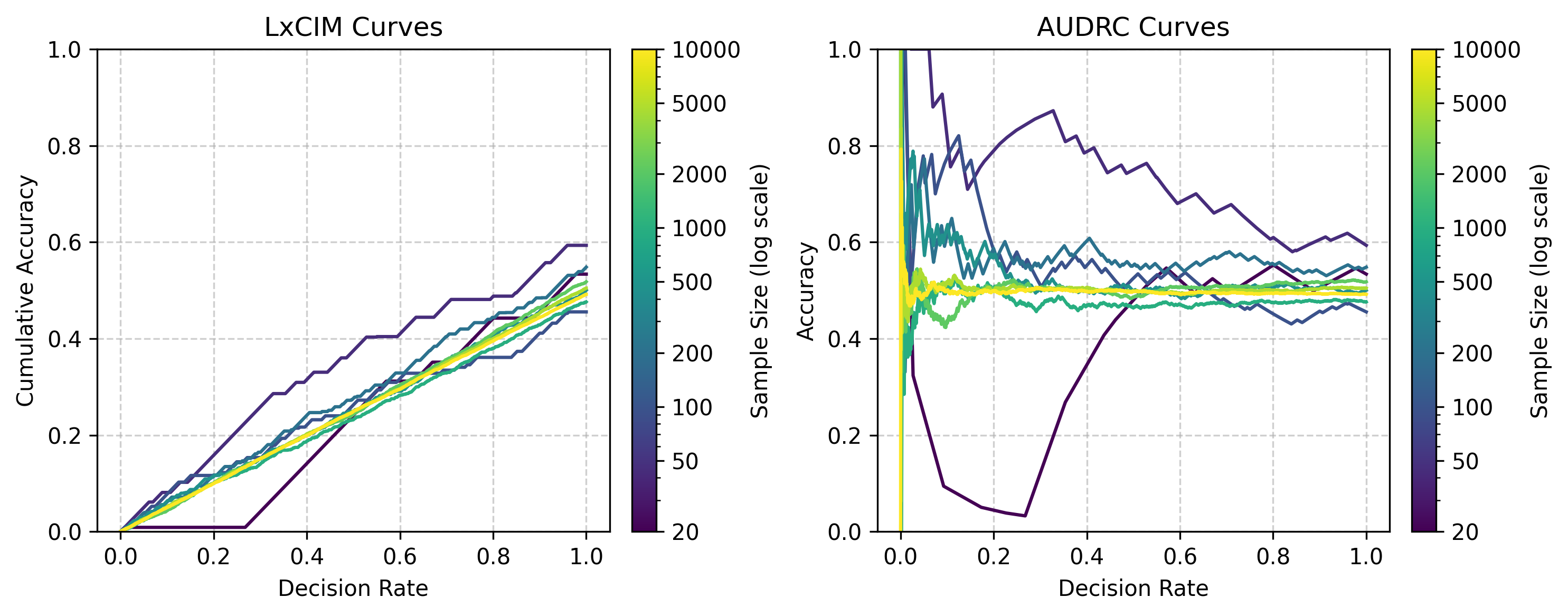}
    \caption{Comparison of the cumulative accuracy-decision rate curve (used for computing LxCIM) and the accuracy-decision rate curve (used for computing AUDRC). Both metrics are evaluated on identical random functions across varying sample sizes, with each sample assigned a uniform weight.}
    \label{random_curves}
\end{figure}

The final major advantage of LxCIM over AUDRC pertains to its stronger mathematical coherence, in specific, its integral remains well-defined and computable even in the limit $N \to \infty$. To illustrate this, Figure \ref{random_curves} plots both LxCIM and AUDRC curves for a purely random classifier evaluated on varying dataset sizes. It is evident that, as $N$ increases, the LxCIM curve smooths naturally, whereas the AUDRC curve remain irregular, particularly at low decision rates, where individual correct or incorrect predictions have a larger effect on the computed accuracy.

Therefore, the non-differentiable nature of AUDRC in the limit $N \to \infty$ occurs because the spacing between successive computable decision rates approaches zero, eliminating the need for interpolation. Consequently, the discrete jumps in accuracy between consecutive predictions manifest as discontinuities in the function, rendering AUDRC non-integrable, as described by Proposition \ref{prop:non_integrable} that is proven in Appendix \ref{non_integrable}.

\begin{proposition}\label{prop:non_integrable}
    The function $\mathrm{AUDRC}$ is not integrable in the limit $N \to \infty$, where $N$ denotes the number of evaluation examples.
\end{proposition}

\section{LxC-invariant binary classification example: bivariate causal discovery}\label{causalexample}

Bivariate causal discovery constitutes a binary classification task aimed at inferring the causal direction between two statistical variables (generally named $X$ and $Y$) solely from observational data \citep{tuebingenresults}. Method performance is generally evaluated on T\"ubingen dataset, which is the only state-of-the-art real-world benchmark in this domain.

Within the bivariate causal discovery literature, the only two dominant binary classification metrics are AUROC and accuracy. While accuracy provides a simple measure of correctness, AUROC is generally favoured as it leverages the rank-based information encoded in the model’s scores, capturing more nuanced distinctions in prediction quality that accuracy alone cannot reflect \citep{tuebingen}. Additionally, \citet{tuebingenresults} propose Bi-AUROC, defined as the average of the AUROC computed first for the class $X \to Y$ versus all other cases, and then for $Y \to X$ versus all others. Despite its potentially misleading name, Bi-AUROC is not a binary classification metric. Rather, it is designed specifically for the bivariate causal discovery problem in contexts where alternative scenarios, such as independence or the presence of confounders, are also considered. 

Nevertheless, \citet{tuebingenresults} overlook the inherent arbitrariness in the labelling process of this pairwise choice problem, evaluating methods solely against the ground-truth directions originally specified in the dataset. In contrast, \citet{tuebingen} explicitly acknowledge this arbitrariness, noting that a key limitation of ROC-based evaluation is the introduction of an asymmetry between positives and negatives. Consequently, they argue that accuracy provides a more natural and unbiased measure. To address this, they propose balancing class labels by swapping $X$ and $Y$ for a subset of the pairs, ensuring symmetry in evaluation. In fact, this issue arises precisely because AUROC is not LxC-invariant, meaning that the arbitrary labelling process directly influences the metric’s outcome.

More recently, \citet{audrc} address the inherent arbitrariness in the labelling process by proposing the use of AUDRC as an alternative rank-based evaluation metric. The use of accuracy-decision rate curves had also already been used to facilitate a deeper analysis in some methods (e.g., bQCD \citep{bcqd}). Nevertheless, as previously noted, \citet{tuebingen} caution against the use of such curves, due to their overemphasis on high-confidence predictions.

In turn, LxCIM directly addresses each of these limitations identified in the evaluation of bivariate causal discovery methods. Unlike accuracy, it incorporates the full spectrum of prediction scores rather than merely considering correctness. Unlike AUROC, it is invariant to local exchange of classes (LxC), rendering it robust to the arbitrary labeling inherent in this pairwise choice problem. And, unlike AUDRC, it attributes linearly more weight to higher confidence guesses, ensuring a fair evaluation across all predictions.

\begin{figure}[h!]
  \centering
  \includegraphics[width=0.55\linewidth]{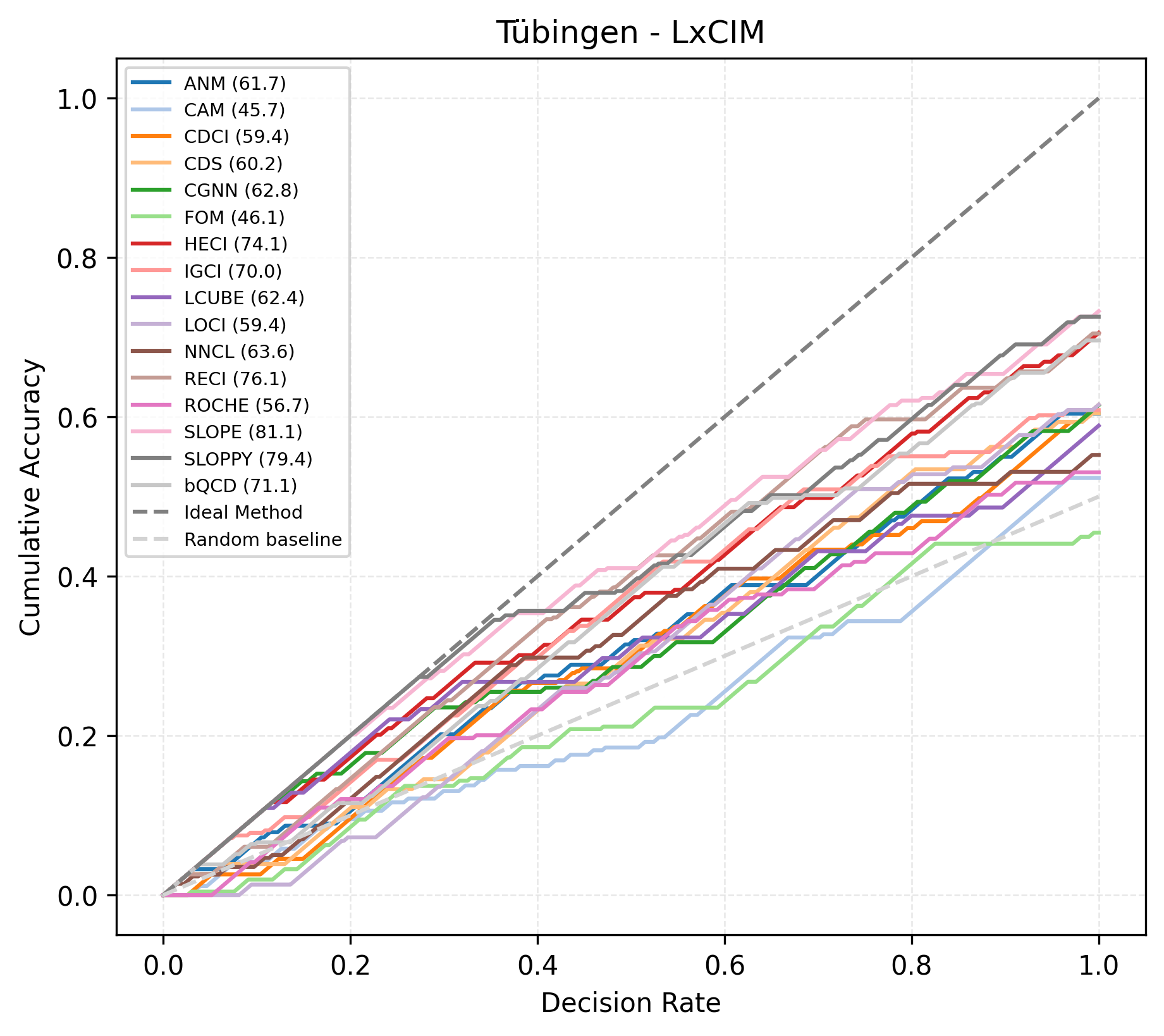}
  \caption{Cumulative accuracy-decision rate curves for all sixteen causal discovery methods on the Tübingen benchmark. Ideal and random curves are also shown for context.}
  \label{alamedacurvestuebingen}
\end{figure}

Having defined the advantages of LxCIM compared to existing state-of-the-art metrics, the performance of 16 bivariate causal discovery methods is evaluated on the T\"ubingen dataset using LxCIM, accuracy, AUROC, and AUDRC. The 16 implemented methods are ANM \citep{peters2010identifying}, bQCD \citep{bcqd}, CAM \citep{cam}, CDCI \citep{cdci}, CDS \citep{cds}, CGNN \citep{cgnn}, FOM \citep{fom}, HECI \citep{heci}, IGCI \citep{mian2023information}, LCUBE \citep{lcube}, LOCI \citep{audrc}, NNCL \citep{nncl}, RECI \citep{reci}, ROCHE \citep{roche}, SLOPE \citep{slope} and SLOPPY \citep{sloppy}\footnote{For bQCD, CAM, CDCI, FOM, HECI, LCUBE, LOCI, NNCL, ROCHE, SLOPE, and SLOPPY, the original implementations provided by the authors were used. For ANM, CDS, CGNN, IGCI, and RECI, the code from the CausalDiscoveryToolbox of \citet{cdtoolbox} was used. Finally, since the original implementation of CGNN was too slow, the hyperparameters were adapted to ensure no example took longer than $30$ seconds to score.}. The overall quantitative results are summarized in Table \ref{tuebingen_benchmark_table}, with the corresponding cumulative accuracy-decision rate curves shown in Figure \ref{alamedacurvestuebingen}.

\begin{table}[ht]
\centering
\small
\begin{tabular}{lllll}
\toprule
Method & \textbf{LxCIM (\%)} & Accuracy (\%) & AUROC (\%) & AUDRC (\%) \\
\midrule
ANM & \textbf{61.7} & 60.4 & 61.1 & 62.9 \\
bQCD & \textbf{71.1} & 69.6 & 70.8 & 70.1 \\
CAM & \textbf{45.7} & 52.3 & 46.5 & 43.1 \\
CDCI & \textbf{59.4} & 61.5 & 61.1 & 55.0 \\
CDS & \textbf{60.2} & 60.5 & 58.6 & 57.8 \\
CGNN & \textbf{62.8} & 61.5 & 63.6 & 69.6 \\
FOM & \textbf{46.1} & 45.5 & 46.1 & 40.9 \\
HECI & \textbf{74.1} & 70.5 & 73.6 & 79.2 \\
IGCI & \textbf{70.0} & 60.9 & 68.1 & 73.3 \\
LCUBE & \textbf{62.4} & 58.9 & 62.2 & 70.1 \\
LOCI & \textbf{59.4} & 61.5 & 58.6 & 49.5 \\
NNCL & \textbf{63.6} & 55.2 & 64.0 & 63.4 \\
RECI & \textbf{76.1} & 70.5 & 77.3 & 75.7 \\
ROCHE & \textbf{56.7} & 53.0 & 58.0 & 53.7 \\
SLOPE & \textbf{\underline{81.1}} & \underline{73.3} & \underline{82.0} & \underline{86.4} \\
SLOPPY & \textbf{79.4} & 72.6 & 78.9 & 85.3 \\
\bottomrule
\end{tabular}

\caption{Performance of each causal discovery method on the Tübingen benchmark. Columns report LxCIM, accuracy, AUROC, and AUDRC for all sixteen implemented methods.}
\label{tuebingen_benchmark_table}
\end{table}

In Table \ref{tuebingen_benchmark_table}, all methods exhibit slightly lower performance when evaluated using accuracy, when compared to AUROC and LxCIM. This discrepancy arises from the fundamental difference in how the metrics treat predictions: accuracy considers only whether each prediction is correct, whereas both AUROC and LxCIM assign greater weight to predictions made with higher confidence.

The second key observation is that AUROC and LxCIM yield very similar performance evaluations across all methods. This outcome is consistent with expectations: as discussed in subsection \ref{auroc_relantionship}, the primary distinction between LxCIM and AUROC is that LxCIM enforces symmetry in the evaluated system, whereas AUROC remains sensitive to asymmetries. Given that the T\"ubingen dataset is comprised of $108$ examples, the effect of the labelling asymmetry is minor, typically impacting the results only in the decimal range, with the notable exceptions of CDCI and CDS, exhibiting a difference of around $2\%$ between the two metrics.

On the other hand, the analysis of the cumulative accuracy as a function of decision rate depicted in Figure \ref{alamedacurvestuebingen} reveals another critical insight: while the score generally serves as a reliable predictor of class, it provides little information regarding each method's confidence in its predictions. This observation is supported by the relatively flat slopes observed for most methods as decision rate increases, indicating that their predictive performance does not degrade substantially even as confidence in individual predictions decreases.

\begin{figure}[h!]
    \centering
    \includegraphics[width=\linewidth]{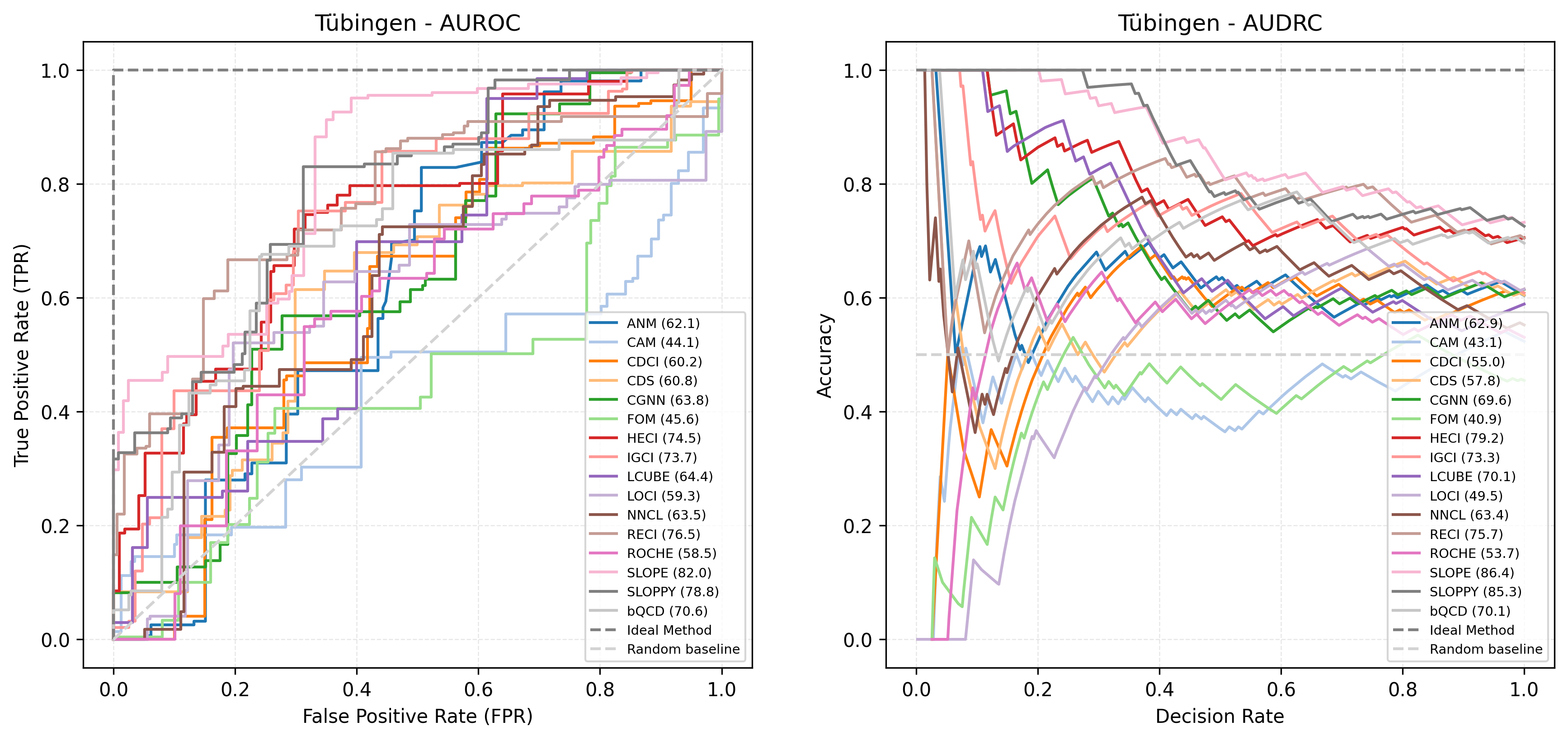}
    \caption{On the left, the ROC curve is plotted for each of the sixteen implemented bivariate causal discovery methods, while on the right, the AUDRC curves are shown.}
    \label{aurocauacctuebingen}
\end{figure}

Finally, a comparison between the cumulative accuracy-decision rate curves present in Figure \ref{alamedacurvestuebingen} and the ROC and accuracy-decision rate curves in Figure \ref{aurocauacctuebingen} highlights the limitations of the latter two. In the ROC curves, the evolution of method performance as a function of confidence is opaque. For instance, the performance degradation of NNCL beyond a $60\%$ decision rate is not noticeable. Additionally, ROC curves can yield misleading interpretations in LxC-invariant tasks, as exemplified by bQCD and IGCI, performing similarly despite significantly separated ROC curves. Lastly, in the accuracy-decision rate plot, the previously discussed over-emphasis on higher-confidence samples becomes apparent: for example, CDCI appears worse than CDS, even though it achieves higher overall accuracy and similar LxCIM, as reported in Table \ref{tuebingen_benchmark_table}.

\section{Conclusion}

This work begins by introducing a new notion of binary classification metric invariance, called local exchange of classes (LxC). This form of metric invariance is particularly relevant in two scenarios: (i) tasks where class biases are permissible and may be learned by the model, and (ii) problems where the evaluated model is intrinsically symmetric, such that any observed class asymmetry is purely random and should not influence the evaluation. We then propose a new rank-based metric that is LxC-invariant, thus called LxCIM.

LxCIM exhibits strong theoretical and practical connections to other established metrics: as accuracy, it is LxC-invariant; it coincides with AUROC for symmetric prediction scenarios; it is closely related to AUDRC. Moreover, by being rank-based, enforcing symmetry between positive and negative classes, assigning importance linearly proportional to confidence, and providing a highly informative cumulative accuracy–decision rate curve, LxCIM improves upon each of these metrics in the evaluation of LxC-invariant problems.

To illustrate LxCIM's advantages, bivariate causal discovery is considered as a canonical LxC-invariant example. In this task, LxCIM addresses several limitations previously identified in the literature regarding AUROC, accuracy, and AUDRC. Furthermore, the evaluation of 16 state-of-the-art causal discovery methods clearly demonstrates the practical advantages of LxCIM in LxC-invariant tasks.

Finally, LxCIM can also be applied to tasks beyond LxC-invariant problems, where class labels carry intrinsic meaning. In order to leverage the additional benefits of LxCIM, such as the interpretability of the cumulative accuracy–decision rate curve, it suffices to adjust the weights of individual examples to account for the class-specific importance, thereby balancing the contributions of type-I and type-II errors, while preserving the metric’s core properties.


\appendix

\section{Proof Proposition \ref{prop:bimono}} \label{decision_function_appendix}
\begin{proof}
Let \(s:\mathcal{X}\to\mathbb{R}\) be any scoring function satisfying the following ordering property: for any two samples \(X_1, X_2 \in \mathcal{X}\),
\begin{equation}
s(X_1) > s(X_2) \quad \Longleftrightarrow \quad 
P[X_1 \in \mathcal{X}^+] > P[X_2 \in \mathcal{X}^+],
\end{equation}
where $\mathcal{X}^+$ is the set of positive samples. Any confidence function $g:\mathbb{R}\to\mathbb{R}$ consistent with this ordering must satisfy
\begin{equation}
\forall s_i \in \mathbb{R}:\ (\forall s < s_i: g(s) > g(s_i)) \quad \text{or} \quad (\forall s > s_i: g(s) > g(s_i)).
\end{equation}
This condition ensures that, if a particular score \(s_i\) is representative of a positive instance, then all larger scores correspond to even greater confidence in membership to $\mathcal{X}^+$, i.e., \(\forall s > s_i, g(s) > g(s_i)\). Conversely, if \(s_i\) corresponds to a negative instance, then all smaller scores correspond to higher confidence in membership to $\mathcal{X}^-$, i.e., $\forall s < s_i, \; g(s) > g(s_i)$.

This structure imposes a bi-monotonic shape on the confidence function: it must increase in at least one direction from every point \(s_i\). Consequently, there must exist a unique point \(s^*\) such that
$g(s)$ is strictly decreasing for $ s < s^*$, and strictly increasing for $ s > s^*$.
\end{proof}

\section{Proof of Proposition \ref{prop:lxc}} \label{equivalence_proof}

\begin{proof}
The equivalence is proved by considering the two directions of implication. To keep the notation shorter, we define the vectors ${\bf c} = [TP,\, FP, \, FN, \, TN]^T$ and ${\bf d} = [TP+TN,\, FP+FN]^T$, which are related via ${\bf d} = {\bf A \, c}$, where
\[
{\bf A} = \begin{bmatrix}
    1 & 0 & 1 & 0 \\ 0 & 1 & 0 & 1
\end{bmatrix}.
\]

Consider first that there is some function $g:\mathbb{R}^2 \rightarrow \mathbb{R}$ such that Equation \ref{eq:lxc} holds. This equality can be written as $f({\bf c}) = g({\bf Ac})$. Vector $[TP + \delta_1 ,\, FP - \delta_2 , \, FN +\delta_2, \, TN - \delta_1]$ can be written in vector notation as ${\bf c} + {\bf B} \boldsymbol{\delta}$, where 
\[
{\bf B} = \begin{bmatrix}
    1 & 0 \\ 0 & 1 \\ -1 & 0 \\ 0 & -1
\end{bmatrix} \;\;\mbox{and}\;\; \boldsymbol{\delta} = \begin{bmatrix}
\delta_1 \\ \delta_2
\end{bmatrix}.
\]
Since ${\bf AB} = \boldsymbol{0}$, because both columns of ${\bf B}$ are in the null space $\mathcal{N}({\bf A})$, 
\[
f({\bf c} +  {\bf B} \boldsymbol{\delta}) = g\bigl( {\bf A}({\bf c} +  {\bf B} \boldsymbol{\delta})\bigr) = g({\bf Ac}) = f({\bf c}),
\]
proving one of the implications. For the reverse implication, assume that it holds that, for any $\boldsymbol{\delta}\in\mathbb{R}^2$, $f({\bf c} + {\bf B} \boldsymbol{\delta}) = f({\bf c})$, and observe that any vector ${\bf c}$ can be decomposed as 
\[
{\bf c} = {\bf c}^{\parallel} + {\bf c}^{\perp}, \;\; \mbox{where} \;\; {\bf c}^{\parallel} = \frac{1}{2} \begin{bmatrix}
    c_1 + c_3 \\ c_2 + c_4 \\ c_1 + c_3 \\ c_2 + c_4
\end{bmatrix}  \;\; \mbox{and} \;\; 
{\bf c}^{\perp} = \frac{1}{2} \begin{bmatrix}
    c_1 - c_3 \\ c_2 - c_4 \\ c_3 - c_1 \\ c_4 - c_2
\end{bmatrix},
\]
where ${\bf c}^{\perp} \in  \mathcal{N}({\bf A})$ and ${\bf c}^{\parallel}$ is in the orthogonal complement. Thus, 
\[
f({\bf c}) = f({\bf c} + {\bf B} \boldsymbol{\delta}) = f( {\bf c}^{\parallel} + {\bf c}^{\perp} + {\bf B}\boldsymbol{\delta} ) = 
f\bigl( {\bf c}^{\parallel} + {\bf B} (\boldsymbol{\delta} + \boldsymbol{\gamma}) \bigr) = f({\bf c}^{\parallel}),
\]
because ${\bf c}^{\perp} = {\bf B} \boldsymbol{\gamma}$, for some $\boldsymbol{\gamma} \in \mathbb{R}^2$. Finally, because ${\bf c}^{\parallel}$ simply contains two copies of ${\bf Ac}$, it results that there must exist some function $g:\mathbb{R}^2\rightarrow \mathbb{R}$ such that $f({\bf c}) = f({\bf c}^{\parallel}) = g({\bf Ac})$, which completes the proof. 

\end{proof}

\section{Proof of Proposition \ref{prop:crossingpoint}}\label{symmetry_axis_proof}

\begin{proof}
Let $\text{TP}$, $\text{TN}$, $\text{FP}$, and $\text{FN}$ denote the true positives, true negatives, false positives, and false negatives, respectively, and let $N = \text{TP} + \text{TN} + \text{FP} + \text{FN}$. Recall:
\begin{equation}
    \text{TPR} = \frac{\text{TP}}{\text{TP} + \text{FN}}, \quad 
    \text{FPR} = \frac{\text{FP}}{\text{FP} + \text{TN}}, \quad 
    \text{TNR} = \frac{\text{TN}}{\text{TN} + \text{FP}}, \quad 
    \text{ACC} = \frac{\text{TP} + \text{TN}}{N}.
\end{equation}
Given the conditions $\text{TPR} = 1 - \text{FPR}$ and $\text{TNR} = 1 - \text{FNR}$, it follows that
\begin{equation}
    \frac{\text{TP}}{\text{TP} + \text{FN}} = 1 - \frac{\text{FP}}{\text{FP} + \text{TN}} = \frac{\text{TN}}{\text{FP} + \text{TN}},
\end{equation}
hence $\text{TPR} = \text{TNR}$.  
Accuracy can be expressed as
\begin{equation}
    \text{ACC} = P(X^+) \cdot \text{TPR} + P(X^-) \cdot \text{TNR},
\end{equation}
where $P(X^+) = \frac{\text{TP} + \text{FN}}{N}$ and $P(X^-) = 1 - P(X^+)$.  
Substituting $\text{TPR} = \text{TNR}$ yields
\begin{equation}
    \text{ACC} = P(X^+) \cdot \text{TPR} + (1 - P(X^+)) \cdot \text{TPR} = \text{TPR}.
\end{equation}
\end{proof}

\section{Proof Proposition \ref{prop:doubling}}\label{doublingproof}

\begin{proof}
Consider $w$ to be naturally ordered according to descending scores. $\text{TPR}$ and $\text{FPR}$ can be defined as functions of the decision rate ($i$) as:
\begin{equation}\label{tprfpr}
\begin{aligned}
    \text{TPR}(i) &= \frac{\int_{0}^{i} \mathbb{I}\!\left[ y_{j(w)}' = y_+ \right] dw}{\int_{0}^{i} \mathbb{I}\!\left[ y_{j(w)}' = y_+ \right] dw + \int_{i}^{2 \sum_k w_k} \mathbb{I}\!\left[ y_{j(w)}' = y_+ \right] dw}, \\
    \text{FPR}(i) &= \frac{\int_{0}^{i} \mathbb{I}\!\left[ y_{j(w)}' = y_- \right] dw}{\int_{0}^{i} \mathbb{I}\!\left[ y_{j(w)}' = y_- \right] dw + \int_{i}^{2 \sum_k w_k} \mathbb{I}\!\left[ y_{j(w)}' = y_- \right] dw}.
\end{aligned}
\end{equation}
\noindent where,
\begin{equation}
    \int_{0}^{i} \mathbb{I}\!\left[ y_{j(w)} = y_+ \right] dw + \int_{i}^{2 \sum_k w_k} \mathbb{I}\!\left[ y_{j(w)} = y_+ \right] dw
    = \int_{0}^{2 \sum_k w_k} \mathbb{I}\!\left[ y_{j(w)} = y_+ \right] dw
    = w_{y_+} = \sum_k w_k
\end{equation}
\noindent and equivalently for $\text{FPR}(i)$.

This is true since the augmentation process duplicates all pairs while flipping the labels, guaranteeing perfectly balanced classes.

Without loss of generality, let $\{y_+, y_-\}$ be defined as $\{1, 0\}$. Then, from Equation~\eqref{A_eq}, the augmented labels are:
\begin{equation}
    y_{j(w)}' = 
    \begin{cases}
        y_{j(w)}, & \text{if } u(s_{j(w)}) = y_+ = 1, \\
        1 - y_{j(w)}, & \text{otherwise.}
    \end{cases}
\end{equation}
Thus,
\begin{equation}\label{minusclass_eq}
\begin{aligned}
    G(i) &= \int_{0}^{i} \; \frac{|u(s_{j(w)}) - y_{j(w)}|}{2}\; dw = \int_{0}^{i} \; y_{j(w)}' \; dw, \\
    \text{TPR}(i) &= \frac{\int_{0}^{i} \; y_{j(w)}' \; dw}{\sum_k w_k}, \\
    \text{FPR}(i) &= \frac{\int_{0}^{i}\; (1 - y_{j(w)}') \; dw}{\sum_k w_k}.
\end{aligned}
\end{equation}

For simplicity\footnote{This was done to clarify reading, but the proof still holds for any value of $\sum_k w_k$.}, and in accordance with good practice, let $\sum_k w_k = 1$.

From the definition of $\text{AUROC}$, one can define:
\begin{equation}\label{Hdef2}
    H = \int_{0}^{1 - \text{ACC}} \text{TPR}(\text{FPR}) \, d\text{FPR}
\end{equation}
\noindent where the upper end of the integral is $1 - \text{ACC}$, since it has been shown that the ROC curve crosses the line $\text{TPR} = 1 - \text{FPR}$ at $\text{FPR} = 1 - \text{ACC}$.

Joining Equations~\eqref{minusclass_eq} and \eqref{Hdef2}, and noticing the equality between the numerator of $\text{TPR}$ and the $G(i)$ function of Equation~\eqref{alameda_eq}:
\begin{equation}\label{H_def}
    H = \int_{0}^{\text{FPR}^{-1}(1 - \text{ACC})} \text{TPR}(i) \frac{d\text{FPR}(i)}{di} \, di
    = \int_{0}^{1} G(i) (1 - y_{j(w)}') \, di
\end{equation}

On the other hand:
\begin{equation}
\label{eq:alameda_1}
\text{LxCIM} = 2 \int_{0}^{1} G(i) \, di
\end{equation}
\begin{equation}
\label{eq:alameda_2}
\begin{aligned}
&= 2 \left( \int_{0}^{1} G(i) y_{j_i}' \, di + \int_{0}^{1} G(i) (1 - y_{j_i}') \, di \right)
\end{aligned}
\end{equation}
where (from Equation~\ref{minusclass_eq}) $G(i) = \int_{0}^{i} y_{j(w)}' \, dw$:
\begin{equation}
\label{eq:alameda_3}
\begin{aligned}
&= 2 \left( \int_{0}^{1} \left( \int_{0}^{i} y_{j(w)}' \, dw \right) y_{j_i}' \, di + \int_{0}^{1} G(i) (1 - y_{j_i}') \, di \right)
\end{aligned}
\end{equation}
The first integral has the form $\int F'(s) F(s) \, ds$, whose solution is:
\begin{equation}
\label{eq:alameda_4}
\begin{aligned}
&= 2 \left( \frac{1}{2} \left( G(1)^2 - G(0)^2 \right) + \int_{0}^{1} G(i) (1 - y_{j_i}') \, di \right) \\
&= 2 \left( \frac{\text{ACC}^2}{2} + \int_{0}^{1} G(i) (1 - y_{j_i}') \, di \right)
\end{aligned}
\end{equation}
Joining with $H$ definition from Equation~\eqref{H_def}:
\begin{equation}
\label{eq:alameda_5}
\begin{aligned}
&= \text{ACC}^2 + 2 \int_{0}^{1} G(i) (1 - y_{j_i}') \, di \\
&= \text{ACC}^2 + 2H
\end{aligned}
\end{equation}
\end{proof}

\section{Proof of Proposition \ref{prop:non_integrable}} \label{non_integrable}

\begin{proof} For simplicity, and without loss of generality, let's consider uniform weights.
In doing so, Equation \eqref{AUACC} defining AUDRC can be rewritten as
\begin{equation}    
\text{AUDRC} =\int_{0}^1 \acc(j) \; d\,\cov(j), 
\label{AUDRC_2}
\end{equation}
\noindent where 
\begin{equation}
\label{eq:acc_cov_j2}
acc(j) = \frac{\sum_{i=1}^{j} \mathbb{I}[u(s_{\pi(i)}) = y_{\pi(i)}]}{j}, \quad
cov(j) = \frac{j}{N}.
\end{equation}
with $(s_i, y_i, w_i)$ denoting the predicted score, true label, and corresponding weight of each example, and $u(s),g(s), \text{and } \pi$ taking the same definitions of Equation \eqref{AUDRC}.

In turn, $acc$ can be written recursively for $j = 1, \dots, N$:
\begin{equation}
\label{eq:acc_recursion}
acc(j+1) =
\begin{cases}
\frac{j}{j+1} \, acc(j), & \text{if } u(s_{\pi(j+1)}) \neq y_{\pi(j+1)},\\[1mm]
\frac{j\, acc(j) + 1}{j+1}, & \text{if } u(s_{\pi(j+1)}) = y_{\pi(j+1)}.
\end{cases}
\end{equation}
The expression above can be rewritten as a function of the decision rate ($i$), where $i=j/N$ and $\delta = 1/N$ as:
\begin{equation}
acc(i+\delta) =
\begin{cases}
\frac{i}{i+\delta} acc(i), & \text{if } u(s_{\pi(\lfloor(i+\delta)*N)\rfloor}) \neq y_{\pi(\lfloor(i+\delta)*N)\rfloor)}\\[1mm]
\frac{i \, acc(i) + 1}{i+\delta}, & \text{if } u(s_{\pi(\lfloor(i+\delta)*N)\rfloor)}) = y_{\pi(\lfloor(i+\delta)*N)\rfloor)}.
\end{cases}
\end{equation}
As $N \to \infty$ and $\delta \to 0$, the derivative does not converge:
\begin{equation}
\begin{aligned}
\lim_{\delta \to 0} 
\frac{acc(i+\delta) - acc(i)}{\delta}
\Big|_{u(s_{\pi(\lfloor (i+\delta)N \rfloor)}) \neq y_{\pi(\lfloor (i+\delta)N \rfloor)}} 
&= 
\frac{\left(\frac{i}{i+\delta} - 1\right) acc_i}{\delta}
= \frac{-\delta \, acc_i}{i\delta + \delta^2}
= -\infty, \\[6pt]
\lim_{\delta \to 0} 
\frac{acc(i+\delta) - acc(i)}{\delta}
\Big|_{u(s_{\pi(\lfloor (i+\delta)N \rfloor)}) = y_{\pi(\lfloor (i+\delta)N \rfloor)}} 
&= 
\frac{\frac{i \, acc_i + 1}{i+\delta} - acc_i}{\delta}
= \frac{1 - \delta \, acc_i}{i\delta + \delta^2}
= +\infty.
\end{aligned}
\end{equation}
Since the derivative is not defined for any 
$i\in\mathbb{R}$, the Riemann integral is not defined.
\end{proof}

\end{document}